\documentclass[10pt,twocolumn,letterpaper]{article}

\usepackage{iccv}
\usepackage{times}
\usepackage{epsfig}
\usepackage{graphicx}
\usepackage{amsmath}
\usepackage{amssymb}
\usepackage{booktabs}
\usepackage{multirow}
\usepackage{enumitem}
\usepackage{amsthm}
\usepackage{algpseudocode, algorithm}

\usepackage[pagebackref=true,breaklinks=true,letterpaper=true,colorlinks,bookmarks=false]{hyperref}

\usepackage[textsize=tiny]{todonotes}

\newtheorem{theorem}{Theorem}[section]

\newtheorem{proposition}[theorem]{Proposition}
\theoremstyle{definition}
\newtheorem{definition}{Definition}[section]

\newtheorem{claim}{Claim}[section]

\newtheorem*{remark}{Remark}

\newcommand{\bftheta}{\boldsymbol{\theta}}
\newcommand{\bfepsilon}{\boldsymbol{\epsilon}}
\newcommand{\bfDelta}{\boldsymbol{\Delta}}

\newcommand{\bfg}{\boldsymbol{g}}
\newcommand{\bfh}{\boldsymbol{h}}
\newcommand{\bbR}{\mathbb{R}}
\newcommand{\bbE}{\mathbb{E}}

\newcommand{\caltheta}{\Theta}
\newcommand{\calD}{\mathcal{D}}
\newcommand{\calX}{\mathcal{X}}
\newcommand{\calY}{\mathcal{Y}}
\newcommand{\calL}{\mathcal{L}}
\newcommand{\allloss}{\mathcal{L}_{\text{all}}}

\newcommand{\gam}{R^{(1)}}
\newcommand{\sam}{R^{(0)}}
\newcommand{\fad}{R^{\text{FAD}}}
\newcommand{\train}{{\text{tr}}}
\newcommand{\test}{{\text{te}}}
\newcommand{\error}{\mathcal{E}}
\newcommand{\calE}{\mathcal{E}}
\newcommand{\errortrain}{\hat{\error}_{\train}}

\newcommand\blfootnote[1]{%
  \begingroup
  \renewcommand\thefootnote{}\footnote{#1}%
  \addtocounter{footnote}{-1}%
  \endgroup
}

\usepackage[capitalize]{cleveref}
\crefname{section}{Sec.}{Secs.}
\Crefname{section}{Section}{Sections}
\Crefname{table}{Table}{Tables}
\crefname{table}{Tab.}{Tabs.}

\def\iccvPaperID{2485} 

\ificcvfinal\fi
\iccvfinalcopy

\makeatletter
\def\maketag@@@#1{\hbox{\m@th\normalfont\normalsize#1}}
\makeatother

\begin{document}

\title{Flatness-Aware Minimization for Domain Generalization}

\author{Xingxuan Zhang, Renzhe Xu, Han Yu, Yancheng Dong, Pengfei Tian, Peng Cui*\\
Department of Computer Science, Tsinghua University\\
{\tt\small {xingxuanzhang@hotmail.com, cuip@tsinghua.edu.cn 
}}}
\maketitle

\begin{abstract}
Domain generalization (DG) seeks to learn robust models that generalize well under unknown distribution shifts. 
As a critical aspect of DG, 
optimizer selection has not been explored in depth. 
Currently, most DG methods follow the widely used benchmark, DomainBed, and utilize Adam as the default optimizer for all datasets. However, we reveal that Adam is not necessarily the optimal choice for the majority of current DG methods and datasets. Based on the perspective of loss landscape flatness, we propose a novel approach, Flatness-Aware Minimization for Domain Generalization (FAD), which can efficiently optimize both zeroth-order and first-order flatness simultaneously for DG. We provide theoretical analyses of the FAD's out-of-distribution (OOD) generalization error and convergence. Our experimental results demonstrate the superiority of FAD on various DG datasets. Additionally, we confirm that FAD is capable of discovering flatter optima in comparison to other zeroth-order and first-order flatness-aware optimization methods.

\end{abstract}

\blfootnote{*Corresponding author}

\section{Introduction}

Current neural networks are expected to generalize to unseen distributions in real-world applications, which can break the independent
and identically distributional (I.I.D.) assumption of traditional machine learning algorithms \cite{zhou2021domain,wang2022generalizing}. The distribution shift between training and test data may largely deteriorate most current approaches in practice. Hence, instead of generalization within the training distribution, the ability to generalize under distribution shift, namely domain generalization (DG) \cite{muandet2013domain}, has attracted increasing attention \cite{wang2022generalizing}.

Various methods have been proposed to address the DG problem recently, many of which have shown promising performance. As a learning problem mainly in the computer vision field, the DG task relies highly on the chosen optimizer, yet the effect of the optimizer in DG has hardly been studied. Most current algorithms, following the well-known benchmark DomainBed \cite{gulrajani2020search}, default to optimize models with Adam \cite{kingma2014adam} without considering other optimizers. 
However, the in-distribution generalization ability of popular optimizers has been investigated from the perspective of loss landscape recently \cite{foret2021sharpness,gan2020large} and Adam is found to achieve inferior generalization compared with other optimizers such as SGD \cite{nesterov1983method} due to the sharp minima Adam selected. Thus, investigating the impact of optimizers in DG and clarifying whether Adam should be considered as the default optimizer is of significance. 

\begin{figure}[t]
    \centering
    \includegraphics[width=\linewidth]{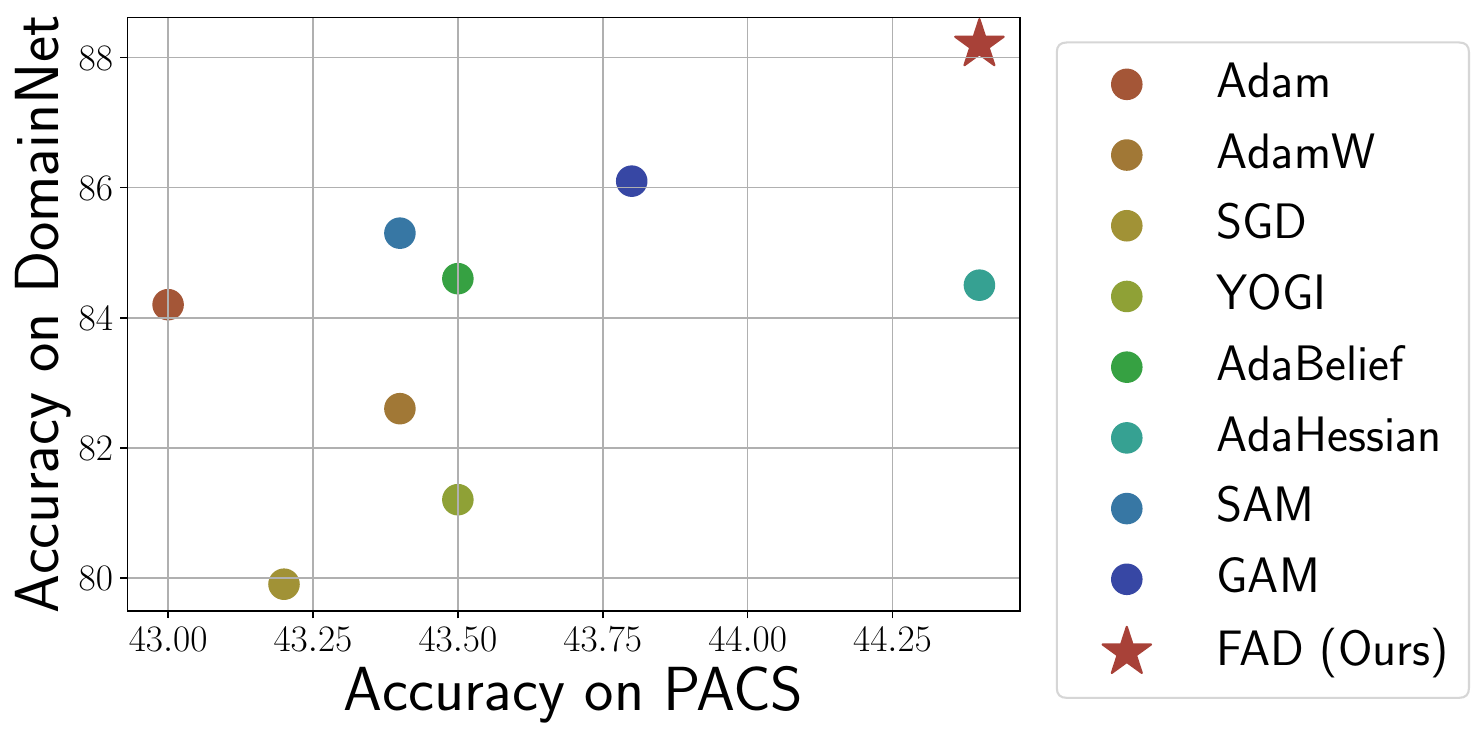}
    \caption{Test accuracy on PACS and DomainNet of different optimizers. Adam, the default optimizer in popular benchmarks shows no advantage compared with its counterparts, while the proposed FAD achieve better OOD generalization performance compared with current optimizers.}
    \label{fig:intro-pacs-domainnet}
\end{figure}

In this paper, we empirically compare the performance of current optimizers, including Adam, SGD, AdaBelief \cite{zhuang2020adabelief}, YOGI \cite{zaheer2018adaptive}, AdaHessian \cite{yao2021adahessian}, SAM \cite{foret2021sharpness}, and GAM\cite{Zhang2023GradientNA}.  
Through extensive experiments on current DG benchmarks, we show that Adam can hardly surpass other optimizers on most DG datasets. The best choice of the optimizer for different DG methods varies across different datasets, and choosing the right optimizer can help exceed the current leaderboards' limitations. 

Recently, the connection between the flatness of the loss landscape and in-distribution generalization ability has been widely studied and verified both theoretically\cite{zhuang2022surrogate} and empirically\cite{du2022sharpness}.  
Some works \cite{cha2021swad, rangwani2022closer} show that flatness also leads to superior OOD generalization. However, none of the previous works consider the optimization in DG nor provide theoretical assurance of their approach. 
Among flatness-aware optimization methods, sharpness-Aware Minimization (SAM) \cite{foret2021sharpness} and its variants \cite{zhuang2022surrogate, kwon2021asam,zhong2022improving,kim2022sharpness}, which optimize the zeroth-order flatness, achieve SOTA performance on various in-distribution image classification tasks \cite{kwon2021asam,zhuang2022surrogate}. Most recently, Gradient Norm Aware Minimization (GAM) \cite{Zhang2023GradientNA} shows that minimizing the first-order flatness provides a more substantial penalty on the sharpness of minima than zeroth-order flatness yet requires more computation overhead. 
Due to the first-order Taylor expansions adopted to optimize zeroth-order and first-order flatness, both SAM and GAM lose some of their penalty strength, and thus they can be mutually reinforced by each other. To accelerate the optimization of first-order flatness and combine zero-order flatness, we propose a unified optimization framework, Flatness-Aware minimization for Domain generalization (FAD), which eliminates the considerable computation overhead for Hessian or Hessian-vector products. 
We theoretically show that the proposed FAD controls the dominant eigenvalue of Hessian of the training loss, which indicates the sharpness of the loss landscape along its most critical direction \cite{kaur2022maximum}.
We present an OOD generalization bound to show that the proposed FAD guarantees the generalization error on test data and convergence analysis. Through extensive experiments, we evaluate FAD and other optimizers on various DG benchmarks.

We summarize the main contribution as follows.
\begin{itemize}[noitemsep,topsep=0pt,parsep=0pt,partopsep=0pt]
    \item We propose a unified optimization framework, Flatness-Aware minimization for Domain generalization (FAD), to optimize zeroth-order and first-order flatness simultaneously. Without calculating Hessian and Hessian-vector products, FAD considerably reduces the computation overhead of first-order flatness.
    \item We theoretically analyze the OOD generalization error and convergence of FAD. We show that FAD controls the dominant eigenvalue of Hessian and thus the flatness of learned minima.
    \item We empirically show that Adam, the default optimizer for most current DG benchmarks, can hardly be the optimal choice for most DG datasets. We present the superiority of FAD compared with current optimizers through extensive experiments.
    \item We empirically validate that FAD finds flatter optima with lower Hessian spectra compared with zeroth-order and first-order flatness-aware optimization methods.
\end{itemize}

\section{Related Works}
\subsection{Optimizers}
Some works \cite{xie2022adaptive,foret2021sharpness} have studied the connection between current optimization approaches, such as SGD \cite{nesterov1983method}, Adam \cite{DBLP:journals/corr/KingmaB14}, AdamW \cite{loshchilov2017decoupled} and others \cite{duchi2011adaptive,DBLP:conf/iclr/LiuJHCLG020} and in-distribution generalization. Some literature shows that Adam is more vulnerable to sharp minima than SGD \cite{wilson2017marginal}, which may result in worse generalization \cite{xie2022power, hardt2016train, hochreiter1994simplifying}. Some follow-ups \cite{luo2019adaptive,chen2018closing,xie2022adaptive,zaheer2018adaptive} propose generalizable optimizers to address this problem. 
However, the generalization ability and convergence speed is often a trade-off \cite{DBLP:conf/iclr/KeskarMNST17,xie2022adaptive,DBLP:conf/iclr/LiuJHCLG020,zaheer2018adaptive,duchi2011adaptive}. 
Different optimizers may favor different tasks and network architectures(e.g., SGD is often chosen for ResNet \cite{he2016deep} while AdamW \cite{loshchilov2017decoupled} for ViTs \cite{dosovitskiy2020image}). Selecting the right optimizer is critical for performance and convergence while the understanding of its relationship to model generalization remains nascent \cite{foret2021sharpness}. 
\cite{naganuma2022empirical} finds that fine-tuned SGD outperforms Adam on OOD tasks. Yet all the previous works do not consider the effect of optimizers in current DG benchmarks and evaluation protocols. In this paper, we discuss the selection of the default optimizer in DG benchmarks.

\subsection{Domain Generalization and Optimization}
Domain generalization (DG) aims to improve the generalization ability to novel domains \cite{ghifary2015domain,khosla2012undoing,koh2021wilds}. 
A common approach is to learn domain-invariant or causal features over multiple source domains \cite{dou2019domain,hu2020domain,li2018learning, li2018domain, motiian2017unified, piratla2020efficient, seo2019learning, zhang2021deep,xu2021stable,zhang2022towards,wang2022causal,zhang2023free}
or aggregating domain-specific modules \cite{mancini2018best, mancini2018robust}. 
Some works propose to enlarge the input space by augmentation of training data \cite{carlucci2019domain, volpi2018generalizing, qiao2020learning, shankar2018generalizing, zhou2020deep, zhou2020learning, Xu2021AFF}. 
Exploit regularization with meta-learning \cite{dou2019domain, li2019episodic} and Invariant Risk Minimization (IRM) framework \cite{arjovsky2019invariant} are also proposed for DG.
Recently, several works propose to ensemble model weights for better generalization \cite{cha2021swad,arpit2021ensemble,rame2022recycling, wortsman2022model,rame2022diverse,lisimple,chu2022dna} and achieve outstanding performance.

\subsection{Flatness of loss landscape}

Recent works study the relationship between flatter minima lead and better generalization on in-distribution data \cite{DBLP:conf/iclr/KeskarMNST17,DBLP:conf/iclr/KeskarMNST17,zhuang2022surrogate,jia2020information,petzka2021relative}.
\cite{kaur2022maximum} reviews the literature related to generalization and the sharpness of minima. It highlights the role of maximum Hessian eigenvalue in deciding the sharpness of minima \cite{DBLP:conf/iclr/KeskarMNST17, wen2019empirical}. 
Some simple strategies are proposed to optimize maximum Hessian eigenvalue, such as choosing a large learning rate \cite{lewkowycz2020large, DBLP:conf/iclr/CohenKLKT21, DBLP:conf/iclr/JastrzebskiSFAT20} and smaller batch size \cite{DBLP:conf/iclr/SmithL18, lewkowycz2020large, jastrzkebski2017three}. 
 
Sharpness-Aware Minimization (SAM) \cite{foret2021sharpness} and its variants \cite{zhuang2022surrogate, kwon2021asam, du2021efficient,liu2022towards, du2022sharpness,mi2022make,zhong2022improving,kim2022sharpness,kim2022fisher}, which are representative training algorithm to seek zeroth-order flatness, show outstanding performance on in-distribution generalization. 
Most recently, several works discuss the relationship between generalization and gradient norm \cite{barrett2020implicit,zhao2022penalizing}. GAM\cite{Zhang2023GradientNA} proposes to optimize first-order flatness for better generalization. \cite{cha2021swad, rangwani2022closer} show that flatness minima also lead to superior OOD generalization. However, none of the previous works consider the optimization in DG nor provide theoretical assurance of their approach. 

\section{Method}
In in-distribution generalization tasks, sharpness-aware optimization methods show outstanding performance \cite{foret2021sharpness}. We theoretically show that optimizing both zeroth-order and first-order flatness strengthens out-of-distribution performance via a generalization bound. We propose a unified optimization framework to optimize zeroth-order and first-order flatness effectively and efficiently. We present the connection between Hessian spectra and our regularization term and give the convergence analysis of our method.

\paragraph{Notations} We use $\calX$ and $\calY$ to denote the space of input $X$ and outcome $Y$, respectively. We use $\Delta_{\calY}$ to denote a distribution on $\calY$.
Let $S = \{(x_i, y_i)\}_{i=1}^n$ denote the training dataset with $n$ data-points drawn independently from the training distribution $\calD_{\train}$. We consider the covariate shift scenario \cite{ben2006analysis,shimodaira2000improving,xu2022theoretical} and assume that the conditional probabilities are invariant across training and test distributions (\textit{i.e.}, $P_{\train}(Y|X) = P_{\test}(Y|X)$). The invariant conditional probability is denoted as $P(Y|X)$.
A prediction model $f_{\bftheta}: \calX \rightarrow \Delta_{\calY}$ parametrized by $\bftheta \in \caltheta \subseteq \mathbb{R}^d$ with dimension $d$ maps an input to a simplex on $\calY$, which indicates the predicted probability of each class\footnote{We use $\Delta_{\calY}$ to denote the non-deterministic function case. This formulation also includes deterministic function cases.}. $\caltheta$ is the hypothesis set. We further assume that the ground truth model $\bftheta^*$ that characterizes $P(Y|X)$ is in the hypothesis set $\caltheta$.

Let $\ell: \Delta_\calY \times \Delta_\calY \rightarrow \bbR_{+}$ define a loss function over $\Delta_\calY$. For any hypotheses $\bftheta_1, \bftheta_2 \in \caltheta$, the expected loss $\calL_{\calD}(\bftheta_1, \bftheta_2)$ for distribution $\calD$ is given as $\calL_{\calD}(\bftheta_1, \bftheta_2) = \bbE_{x \sim \calD}\left[\ell(f_{\bftheta_1}(x), f_{\bftheta_2}(x))\right]$. To simplify the notations, we use $\calL_\train$ and $\calL_\test$ to denote the expected loss $\calL_{\calD_\train}$ and $\calL_{\calD_\test}$ in training and test distributions, respectively. In addition, we use $\error_{\train}(\bftheta) = \calL_{\train}\left(\bftheta, \bftheta^*\right)$ and $\error_{\test}(\bftheta) = \calL_{\test}\left(\bftheta, \bftheta^*\right)$ to denote the loss of a function $\bftheta \in \caltheta$ \textit{w.r.t.} the true labeling function 
$\bftheta^*$ in the two distributions. We use $\errortrain(\bftheta)$ to denote the empirical loss function \textit{w.r.t.} the $n$ samples $S$ from the training distribution. In addition, we assume that $\errortrain(\bftheta)$ is twice differentiable throughout the paper and we use $\nabla \errortrain(\bftheta)$ and $\nabla^2 \errortrain(\bftheta)$ to denote the derivative and Hessian matrix, respectively.

\subsection{A General Flatness-Aware Optimization for DG}

In in-distribution generalization tasks, it is shown that the zeroth-order flatness can be not strong enough in cases where $\rho$ covers multiple minima or the maximum loss in $\rho$ is misaligned with the uptrend of loss and first-order flatness can help improve it \cite{Zhang2023GradientNA}. 
But both zeroth-order and first-order flatness require first-order approximations in practice.
Thus optimizing both zeroth-order and first-order flatness together leads to stronger generalization than both of them \cite{Zhang2023GradientNA}. 
Yet optimizing first-order flatness requires the calculation of Hessian or Hessian-vector products, which introduces heavy computation overhead. As a result, we aim to accelerate the optimization in a Hessian-free way for OOD generalization.

Formally, the zeroth-order and first-order flatness are defined as follows \cite{foret2021sharpness,zhuang2022surrogate,Zhang2023GradientNA}.
\begin{definition}[$\rho$-zeroth-order flatness] \label{defn:zeroth-order}
    For any $\rho > 0$, the $\rho$-zeroth-order flatness $\sam_{\rho}(\bftheta)$ of function $\errortrain(\bftheta)$ at a point $\bftheta$ is defined as
    \begin{equation} \label{eq:sam}
    \small
        \sam_{\rho}(\bftheta) \triangleq \max_{\bftheta' \in B(\bftheta, \rho)} \left(\errortrain(\bftheta') - \errortrain(\bftheta)\right), \quad \forall \bftheta \in \caltheta.
    \end{equation}
    Here $\rho$ is the perturbation radius that controls the magnitude of the neighborhood.
\end{definition}

\begin{definition}[$\rho$-first-order flatness] \label{defn:first-order}
    For any $\rho > 0$, the $\rho$-first-order flatness $\gam_{\rho}(\bftheta)$ of function $\errortrain(\bftheta)$ at a point $\bftheta$ is defined as
    \begin{equation} \label{eq:GAM}
        \gam_{\rho}(\bftheta) \triangleq \rho \cdot \max_{\bftheta' \in B(\bftheta, \rho)} \left\|\nabla \errortrain(\bftheta')\right\|, \quad \forall \bftheta \in \caltheta.
    \end{equation}
    Here $\rho$ is the perturbation radius similar to Definition \ref{defn:zeroth-order}.
\end{definition}

Based on these definitions, we propose to jointly optimize zeroth-order and first-order flatness through the following linear combination of the two flatness metrics.
\begin{equation} \label{eq:fad}
    \fad_{\rho, \alpha}(\bftheta) = \alpha \sam_{\rho}(\bftheta) + (1 - \alpha) \gam_{\rho}(\bftheta),
\end{equation}
where $\alpha$ is a hyper-parameter that controls the trade-off between zeroth-order and first-order flatness. Since both flatness are related to the maximal eigenvalue of the Hessian matrices \cite{zhuang2022surrogate,Zhang2023GradientNA}, we can derive that $\fad_{\rho, \alpha}(\bftheta)$ is also related to the maximal eigenvalue. Specifically, when a point $\bftheta^*$ is a local minimum of $\errortrain(\bftheta)$ and $\errortrain(\bftheta)$ can be second-order Taylor approximated in the neighbourhood of $\bftheta^*$, then
\begin{equation} \label{eq:fad-eigen}
  \lambda_{\max}\left(\nabla^2 \errortrain(\bftheta^*)\right) = \frac{\fad_{\rho, \alpha}(\bftheta^*)}{\rho^2\left(1 - \frac{\alpha}{2}\right)}.
\end{equation}
See Appendix A for details. Since the maximal eigenvalue is a proper measure of the curvature of minima \cite{kaur2022maximum,keskar2016large} and related to generalization abilities \cite{Zhang2023GradientNA,chen2021vision,jastrzkebski2017three}, optimizing Equation \eqref{eq:fad} would hopefully improve the DG performances, which we demonstrate in Section~\ref{sect:fad-and-dg}.

Then the overall objective of our method, Flatness-Aware minimization for Domain generalization (FAD), is as follows.
\begin{equation} \label{eq:all-loss}
    \allloss(\bftheta) = \errortrain(\bftheta) + \beta \cdot \fad_{\rho, \alpha}(\bftheta).
\end{equation}
The strength of FAD regularization is controlled by a hyperparameter $\beta$.

\subsection{Flatness Penalty and DG Performance} \label{sect:fad-and-dg}

We give a generalization bound to demonstrate that FAD controls the generalization error on test distribution as follows. 

\begin{proposition} \label{prop:ood-bound}
    Suppose the per-data-point loss function $\ell$ is differentiable, symmetric, bounded by $M$, and obeys the triangle inequality. Suppose $\bftheta^* \in \caltheta$. Fix $\rho > 0$ and $\bftheta \in \caltheta$. Then with probability at least $1 - \delta$ over training set $S$ generated from the distribution $\calD$,
    \begin{equation} \label{eq:main-bound}
        \small
        \begin{aligned}
            & \, \bbE_{\epsilon_i \sim N(0, \rho^2/(\sqrt{d} + \sqrt{\log n})^2)}[\error_{\test}(\bftheta + \bfepsilon)] \\
            \le & \, \errortrain(\bftheta) + \fad_{\rho, \alpha}(\bftheta) \\ 
            + & \, \sup_{\bftheta_1, \bftheta_2 \in \caltheta}\left|\calL_{\train}(\bftheta_1, \bftheta_2) - \calL_{\test}(\bftheta_1, \bftheta_2)\right| + \frac{M}{\sqrt{n}} \\
            + & \, \sqrt{\frac{\frac{1}{4} d \log \left(1+\frac{\|\bftheta\|^2\left(\sqrt{d} + \sqrt{\log n}\right)^2}{d \rho^2}\right)+\frac{1}{4}+\log \frac{n}{\delta}+2 \log (6 n+3 d)}{n-1}}.
        \end{aligned}
    \end{equation}
\end{proposition}

\begin{remark}
    The term $\sup_{\bftheta_1, \bftheta_2 \in \caltheta}|\cdot|$ corresponds to the discrepancy distance \cite{mansour2009domain}, which measures the covariate shift between the training domain and the unknown test domain. Please note that in DG/OOD generalization tasks, the information on test distribution is unavailable in the optimization phase. Thus one cannot intervene in the third term on RHS. Thus optimizing the regularization of FAD (the second term on RHS) with ERM loss on training data leads to better generalization on test data.
    
\end{remark}

\subsection{Algorithm Details} \label{sect:algorithm-detail}
\begin{algorithm}[t]
\caption{Flatness-Aware Minimization for Domain Generalization (FAD)}
\label{alg:all}
\begin{algorithmic}[1]
    \State \textbf{Input:} Batch size $b$, Learning rate $\eta_t$, Perturbation radius $\rho_t$, Trade-off coefficients $\alpha$, $\beta$, Small constant $\xi$
    \State $t \leftarrow 0$, $\bftheta_0 \leftarrow$ initial parameters
    \While{$\bftheta_t$ not converged}
        \State Sample $W_t$ from the training data with $b$ instances
        \State $\bfg_{t,0} \leftarrow \nabla \hat{\error}_{W_t}(\bftheta_t)$
        \State $\tilde{\bftheta}_{t,1} \leftarrow \bftheta_t + \rho_t \cdot \bfg_{t,0} / (\|\bfg_{t,0}\|+\xi)$
        \State $\bfg_{t,1} \leftarrow \nabla \hat{\error}_{W_t}(\tilde{\bftheta}_{t,1})$
        \State $\bfh_{t,0} \leftarrow \bfg_{t,1} - \bfg_{t,0}$ \Comment{gradient of $\sam_{\rho_t}(\bftheta_t)$}
        \State $\tilde{\bftheta}_{t,2} \leftarrow \bftheta_t + \rho_t \cdot \bfh_{t,0} / (\|\bfh_{t,0}\|+\xi)$
        \State $\bfg_{t,2} \leftarrow \nabla \hat{\error}_{W_t}(\tilde{\bftheta}_{t,2})$
        \State $\tilde{\bftheta}_{t,3} \leftarrow \tilde{\bftheta}_{t,2} + \rho_t \cdot \bfg_{t,2} / (\|\bfg_{t,2}\|+\xi)$
        \State $\bfg_{t,3} = \nabla \hat{\error}_{W_t}(\tilde{\bftheta}_{t,3})$
        \State $\bfh_{t,1} \leftarrow \bfg_{t,3} - \bfg_{t,2}$ \Comment{gradient of $\gam_{\rho_t}(\bftheta_t)$}
        \State $\bftheta_{t+1} \leftarrow \bftheta_t - \eta_t(\bfg_{t,0} + \beta(\alpha \bfh_{t,0} + (1 - \alpha)\bfh_{t,1}))$
        \State $t \leftarrow t + 1$
    \EndWhile
    \State \Return{$\bftheta_t$}
\end{algorithmic}
\end{algorithm}

Although the regularization of FAD controls the OOD generalization error, directly optimizing it 
suffers from significant computational costs.
Previous works \cite{yao2020pyhessian,Zhang2023GradientNA} propose to calculate second order gradient with Hessian-vector product, yet the cost increases fast as the model dimension grows. Inspired by \cite{singh2020woodfisher,zhao2022penalizing}, we approximate second-order gradient with first-order gradients and optimize zeroth-order and first-order flatness simultaneously as follows. 

\paragraph{Optimization and acceleration}
At each round, suppose the perturbation radius is set to $\rho_t$, then the gradient of $\allloss(\bftheta_t)$ at the point $\bftheta_t$ in Equation \eqref{eq:all-loss} is approximated as follows.

\noindent \underline{The gradient of $\sam_{\rho_t}(\bftheta_t)$}. Following \cite{foret2021sharpness}, let
\begin{equation}
    \begin{aligned}
        \tilde{\bftheta}_{t, 1} & = \bftheta_t + \rho_t \cdot \frac{\bfg_{t,0}}{\|\bfg_{t,0}\|} \quad \text{where} \quad \bfg_{t,0} = \nabla \errortrain (\bftheta_t), \\
        \bfg_{t,1} & = \nabla \errortrain(\tilde{\bftheta}_{t,1}).
    \end{aligned}
\end{equation}
Then the gradient of $\sam_{\rho_t}(\bftheta_t)$ is approximated by
\begin{equation} \label{eq:gradient-sam}
    \nabla \sam_{\rho_t}(\bftheta_t) \approx \bfg_{t,1} - \bfg_{t,0}.
\end{equation}

\noindent \underline{The gradient of $\gam_{\rho_t}(\bftheta_t)$}. \cite{Zhang2023GradientNA} propose to approximate the gradient through the following procedure
\begin{equation*}
    \small
    \nabla \gam_{\rho_t}(\bftheta_t) \approx \rho_t \cdot \nabla \left\|\nabla \errortrain(\bftheta^{\text{adv}}_t)\right\|, \bftheta^{\text{adv}}_t = \bftheta_t + \rho_t \cdot \frac{\nabla \left\|\nabla \errortrain(\bftheta_t)\right\|}{\left\|\nabla \|\nabla \errortrain(\bftheta_t)\|\right\|}.
\end{equation*}
One can approximate $\nabla\|\nabla \errortrain(\bftheta_t)\|$ with first-order gradient as follows.
\begin{equation*}
    \nabla\left\|\nabla \errortrain(\bftheta_t)\right\| \approx \frac{\nabla \errortrain(\tilde{\bftheta}_{t,1}) - \nabla \errortrain(\bftheta_t)}{\rho_t} = \frac{\bfg_{t,1} - \bfg_{t,0}}{\rho_t},
\end{equation*}
Thus we can approximate the gradient of $\gam_{\rho_t}(\bftheta_t)$ efficiently and rewrite the term as follows. The details of the derivation can be found in Appendix A.
\begin{equation} \label{eq:gradient-gam}
    \begin{aligned}
        \nabla \gam_{\rho_t}(\bftheta_t) & \approx \bfg_{t,3} - \bfg_{t,2}, \quad \text{where} \\
        \tilde{\bftheta}_{t,2} & = \bftheta_t + \rho_t \cdot \frac{\bfg_{t,1} - \bfg_{t,0}}{\|\bfg_{t,1} - \bfg_{t,0}\|}, \quad \bfg_{t,2} = \nabla \errortrain(\tilde{\bftheta}_{t,2}), \\
        \tilde{\bftheta}_{t,3} & = \tilde{\bftheta}_{t,2} + \rho_t \frac{\bfg_{t,2}}{\|\bfg_{t,2}\|}, \quad \bfg_{t,3} = \nabla \errortrain(\tilde{\bftheta}_{t,3}).
    \end{aligned}
\end{equation}
Please note that the terms $\bfg_{t,1}$ and $\bfg_{t,1}$ for optimizing zeroth-order flatness $\sam_{\rho_t}(\bftheta_t)$ are preserved in the approximation of $\gam_{\rho_t}(\bftheta_t)$ and thus causes no overhead.
Combining the gradient approximation steps as shown in Equations \eqref{eq:gradient-sam} and \eqref{eq:gradient-gam}, we can obtain the unified optimization framework as shown in Algorithm \ref{alg:all}.

\paragraph{Convergence analysis} To analyze the convergence property of Algorithm \ref{alg:all}, we first introduce the Lipschitz smooth assumption that are adopted commonly in optimization-related literature \cite{allen2018neon2,xu2018first,zhuang2022surrogate}.

\begin{definition} \label{defn:loss-smooth}
    For a function $J: \caltheta \rightarrow \bbR$, $J$ is $\gamma_2$-Lipschitz smooth if
    \begin{equation}
        \forall \bftheta_1, \bftheta_2 \in \caltheta, \left\|\nabla J(\bftheta_1) - \nabla J(\bftheta_2)\right\| \le \gamma_2 \|\bftheta_1 - \bftheta_2\|.
    \end{equation}
\end{definition}

Let $\bfDelta_t = \bfg_{t,0} + \beta(\alpha \bfh_{t,0} + (1 - \alpha)\bfh_{t,1})$ be the estimated gradient at round $t$ as shown in Line 14 in Algorithm~\ref{alg:all}. Now we can prove that the algorithm would eventually converge by the following theorem.

\begin{theorem} \label{thrm:convergence-main}
    Suppose $\errortrain(\bftheta)$ is $\gamma$-Lipschitz smooth and bounded by $M$. For any timestamp $t \in \{0, 1, \dots, T\}$ and any $\bftheta \in \caltheta$, suppose we can obtain noisy and bounded observations $f_t(\bftheta)$ of $\nabla \errortrain(\bftheta)$ such that
    \begin{equation}
        \bbE[f_t(\bftheta)] = \nabla \errortrain(\bftheta), \quad \|f_t(\bftheta)\| \le G.
    \end{equation}
    Then with learning rate $\eta_t = \eta_0 / \sqrt{t}$ and perturbation radius $\rho_t = \rho_0 / \sqrt{t}$,
    \begin{equation}
        \sum_{t=1}^T \bbE\left[\left\|\bfDelta_t\right\|^2\right] \le \frac{C_1 + C_2 \log T}{\sqrt{T}},
    \end{equation}
    for some constants $C_1$ and $C_2$ that only depend on $\gamma_1, \gamma_2, G, M, \eta_0, \rho_0$, $\alpha, \beta$.
\end{theorem}

Thus, the convergence of FAD is guaranteed, and we show the convergence rate of FAD empirically in Section~\ref{sec:overhead}.
\section{Experiments}
\begin{table*}[th]
\centering
\caption{Comparison of current optimizers on DG datasets. The best results for each dataset are highlighted in bold font.}
\begin{tabular}{lcccccc|c}
\toprule
\textbf{Algorithm} & PACS & VLCS & OfficeHome & TerraInc & DomainNet & NICO++ & \textbf{Avg.} \\
\midrule

Adam \cite{kingma2014adam} & 84.2$_{\pm 0.6}$ & 77.3$_{\pm 1.3}$ & 67.6$_{\pm 0.4}$ & 44.4$_{\pm 0.8}$ & 43.0$_{\pm 0.1}$ & 76.9$_{\pm 0.0}$ & 65.6 \\
AdamW \cite{loshchilov2017decoupled} & 83.6$_{\pm 1.5}$ & 77.4$_{\pm 0.8}$ & 68.8$_{\pm 0.6}$ & 45.2$_{\pm 1.4}$ & 43.4$_{\pm 0.1}$ & 77.5$_{\pm 0.1}$ & 66.0 \\
SGD \cite{nesterov1983method} & 79.9$_{\pm 1.4}$ & 78.1$_{\pm 0.2}$ & 68.5$_{\pm 0.3}$ & 44.9$_{\pm 1.8}$ & 43.2$_{\pm 0.1}$ & 77.2$_{\pm 0.2}$ & 65.3 \\
YOGI \cite{zaheer2018adaptive} & 81.2$_{\pm 0.4}$ & 77.6$_{\pm 0.6}$ & 68.3$_{\pm 0.3}$ & 45.4$_{\pm 0.5}$ & 43.5$_{\pm 0.0}$ & 77.9$_{\pm 0.2}$ & 65.7 \\
AdaBelief \cite{zhuang2020adabelief} & 84.6$_{\pm 0.6}$ & 78.4$_{\pm 0.4}$ & 68.0$_{\pm 0.9}$ & 45.2$_{\pm 2.0}$ & 43.5$_{\pm 0.1}$ & 77.4$_{\pm 0.1}$ & 66.2 \\
AdaHessian \cite{yao2021adahessian} & 84.5$_{\pm 1.0}$ & 78.6$_{\pm 0.8}$ & 68.4$_{\pm 0.9}$ & 44.4$_{\pm 0.5}$ & \textbf{44.4}$_{\pm 0.1}$ & 77.7$_{\pm 0.1}$ & 66.3 \\
SAM \cite{foret2021sharpness} & 85.3$_{\pm 1.0}$ & 78.2$_{\pm 0.5}$ & 68.0$_{\pm 0.8}$ & \textbf{45.7}$_{\pm 0.9}$ & 43.4$_{\pm 0.1}$ & 77.9$_{\pm 0.1}$ & 66.4 \\
GAM \cite{Zhang2023GradientNA} & 86.1$_{\pm 0.6}$ & 78.5$_{\pm 0.4}$ & 68.2$_{\pm 1.0}$ & 45.2$_{\pm 0.6}$ & 43.8$_{\pm 0.1}$ & 78.0$_{\pm 0.2}$ & 66.6 \\
\midrule
FAD (Ours) & \textbf{88.2}$_{\pm 0.5}$ & \textbf{78.9}$_{\pm 0.8}$  &  \textbf{69.2}$_{\pm }0.5$ & \textbf{45.7}$_{\pm 1.0}$ &  \textbf{44.4}$_{\pm }0.1$ & \textbf{79.0}$_{\pm 0.1}$ & \textbf{67.6} \\
\bottomrule
\end{tabular}
\label{tab:dg-optimizer}
\end{table*}

\subsection{Experimental Settings}
We evaluate our optimization method and current optimizers on various DG benchmarks including PACS \cite{li2017deeper}, VLCS \cite{fang2013unbiased}, OfficeHome \cite{venkateswara2017deep}, TerraIncognita \cite{beery2018recognition}, DomainNet \cite{peng2019moment} and NICO++ \cite{zhang2022nico++}. Detailed introductions of these datasets are in Appendix. We consider the generalization accuracy, training time, and hessian spectra of found minima as the evaluation metrics. We show how optimizers contribute when training with ERM and current SOTA DG algorithms. Generally, Adam is hardly the best choice for DG benchmarks.

For a fair comparison, we follow the basic training and evaluation protocol introduced in \cite{gulrajani2020search}, where the information of test data is unavailable for hyperparameter search. 
We train all the models on DomainNet for 15,000 iterations as suggested in \cite{cha2021swad}, 10,000 iterations on NICO++, and 5,000 iterations on other datasets unless otherwise noted. 
For datasets except for NICO++, we follow the leave-one-out protocol in \cite{gulrajani2020search} where one domain is chosen as the target domain and the remaining domains as the training domain.
For NICO++, we choose two domains as target domains in each run and train models on the remaining four domains. Following the official combination \cite{zhang2022nico++}, we select \{\textit{autumn
}, \textit{rock}\}, \{\textit{dim},\textit{grass}\}, \{\textit{outdoor}, \textit{water}\} as the target domain pairs. A detailed description of the split of training and test domains is in Appendix B. 
For all datasets, 20\% samples in the training data are used for validation and model selection. ImageNet \cite{deng2009imagenet} pretrained ResNet-50 \cite{he2016deep} is adopted as the initial model. The search space for the initial learning rate, weight decay, and other hyperparameters is in Appendix B.

\subsection{Comparison of Optimizers on DG}

Here we investigate how the optimizers affect the generalization under domain shifts. We first conduct experiments with various optimizers following the validation with training data protocol in DomainBed \cite{gulrajani2020search} to study the generalization of current optimizers and the robustness of the choice of hyperparameters. The results are shown in Table~\ref{tab:dg-optimizer}. Detailed results of optimizers on all the splits of reported datasets are in Appendix B.

Despite SGD remaining the most popular optimizer for most CNN-based models on in-distribution image recognition benchmarks, such as CIFAR \cite{krizhevsky2009learning}, and ImageNet due to its outstanding performance, it fails to show advantages compared with other optimizers on current benchmarks for DG. 
The previous work \cite{naganuma2022empirical} shows that SGD with momentum outperforms adaptive-based optimizers such as Adam with  an exhaustive hyperparameter search (hundreds of trials for each optimizer on each dataset) to select hyperparameters with good in-distribution performance. Yet for fair comparisons with previous works, we follow the model selection protocol in DomainBed, where only 20 trials of search are conducted for each evaluation, resulting in inadequate searches. In our experiments, SGD fails to outperform adaptive-based optimizers (e.g., AdamW and AdaBelief). This may be because SGD is more sensitive to the choice of hyperparameters \cite{kingma2014adam} and the hyperparameter search space in evaluation protocols for DG is too large to be traversed and fine-tune the hyperparameters within several random runs. The empirical demonstration of the sensitivity of optimizers to hyperparameters is in Appendix B.

Furthermore, it is important to note that different datasets may require different optimization strategies. For example, Adam outperforms SGD considerably on PACS, while the opposite stands for VLCS, OfficeHome, and NICO++. Among none-flatness-aware optimizers, AdaBelief shows the best average performance. AdamW outperforms Adam on 5 out of 6 datasets and achieves higher average accuracy. Thus, Adam is hardly the best choice for DG and choosing the right optimizer can help exceed current leaderboards' limitations.    

Among all optimizers, flatness-aware optimizers (e.g., SAM, GAM, and FAD) show promising results across all the datasets. FAD considerably outperforms all its counterparts on all the benchmarks, indicating that optimizing zeroth-order and first-order flatness simultaneously learns flatter minima and better generalization.

\subsection{Ensemble with current DG algorithms}
As a base optimizer, FAD can be combined with current DG methods with various objective functions. The compatibility with DG methods is critical for the choice of optimizers.
In this subsection, we investigate the compatibility of GAM with current DG algorithms and compare it with other optimizers. We also report the results of other DG methods for better comparisons.

Please note that we only consider a single pretrained model (i.e., ImageNet pretrained ResNet-50) as the initial model for fair comparisons in this paper, so we do not combine optimizers with some SOTA methods, such as SIMPLE \cite{lisimple}, MIRO \cite{cha2022domain} with RegNetY-16GF \cite{radosavovic2020designing}, and CAR-FT \cite{mao2022context} with CLIP \cite{radford2021learning} which involve the knowledge of pretrained models other than ImageNet pretrained ResNet-50. 
We consider SOTA and representative methods including CORAL \cite{CORAL16}, SWAD \cite{cha2021swad}, and Fishr \cite{rame2022fishr} as the base approach to compare optimizers and leave more methods, including EoA \cite{arpit2021ensemble} and RSC \cite{huang2020self} in Appendix B.  

The results are shown in Table \ref{tab:dg-all}. Different optimizers seem to favor different DG methods. AdamW achieves the best performance when combined with SWAD and CORAL, while SGD surpasses Adam and AdamW with Fishr. This indicates that fixing a single optimizer in DG benchmarks may be unfair for some methods that are not favored, and choosing the proper optimizer can further improve DG methods' performance. 

FAD consistently outperforms other optimizers with all the methods across various datasets. Compared with the default optimizer Adam, FAD further improves SWAD by 1.7\% on PACS and 1.2\% on average and surpasses previous SOTA methods.

\begin{table*}[th]
\centering
\caption{Ensemble with domain generalization methods. Numbers for methods marked with $^*$ and all the combinations of DG methods with optimizers other than Adam are reproduced results. Other results are from the original literature and DomainBed (donated with $^{\dagger}$).}
\resizebox{0.7\linewidth}{!}{
\begin{tabular}{lccccc|c}
\toprule
\textbf{Algorithm} & PACS & VLCS & OfficeHome & TerraInc & DomainNet & \textbf{Avg.} \\
\midrule
MASF \cite{NEURIPS2019_2974788b} & 82.7 & - & - & - & - & - \\
DMC \cite{Chattopadhyay20} & 83.4 & - & - & - & 43.6 & - \\
MetaReg \cite{NEURIPS2018_647bba34} & 83.6 & - & - & - & 43.6 & - \\
ER \cite{NEURIPS2020_b98249b3} & 85.3 & - & - & - & - & - \\
pAdalN \cite{AdaIN21} & 85.4 & - & - & - & - & - \\
EISNet \cite{Wang2020LearningFE} & 85.8 & - & - & - & - & - \\
DSON \cite{Seonguk20} & 86.6 & - & - & - & - & - \\
ERM$^{\dagger}$ \cite{vapnik1999overview} & 85.5 & 77.5 & 66.5 & 46.1 & 40.9 & 63.3 \\
ERM$^*$ (with Adam) & 84.2 & 77.3 & 67.6 & 44.4 & 43.0 & 63.3 \\
IRM$^{\dagger}$ \cite{Arjovsky19} & 83.5 & 78.6 & 64.3 & 47.6 & 33.9 & 61.6 \\
GroupDRO$^{\dagger}$ \cite{Sagawa20Distributionally} & 84.4 & 76.7 & 66.0 & 43.2 & 33.3 & 60.7 \\
I-Mixup$^{\dagger}$ \cite{XuMinghao19,YanShen20,WangYufei20} & 84.6 & 77.4 & 68.1 & 47.9 & 39.2 & 63.4 \\
MLDG$^{\dagger}$ \cite{LiDa17} & 84.9 & 77.2 & 66.8 & 47.8 & 41.2 & 63.6 \\
MMD$^{\dagger}$ \cite{LiHaoliang18} & 84.7 & 77.5 & 66.4 & 42.2 & 23.4 & 58.8 \\
DANN$^{\dagger}$ \cite{GaninYaroslav15} & 83.7 & 78.6 & 65.9 & 46.7 & 38.3 & 62.6 \\
CDANN$^{\dagger}$ \cite{LiYa18} & 82.6 & 77.5 & 65.7 & 45.8 & 38.3 & 62.0 \\
MTL$^{\dagger}$ \cite{BlanchardGilles17} & 84.6 & 77.2 & 66.4 & 45.6 & 40.6 & 62.9 \\
SagNet$^{\dagger}$ \cite{Nam2021ReducingDG} & 86.3 & 77.8 & 68.1 & 48.6 & 40.3 & 64.2 \\
ARM$^{\dagger}$ \cite{zhang2021adaptive} & 85.1 & 77.6 & 64.8 & 45.5 & 35.5 & 61.7 \\
VREx$^{\dagger}$ \cite{KruegerDavid20} & 84.9 & 78.3 & 66.4 & 46.4 & 33.6 & 61.9 \\
RSC$^{\dagger}$ \cite{Huangzeyi20} & 85.2 & 77.1 & 65.5 & 46.6 & 38.9 & 62.7 \\
Mixstyle \cite{zhou2021domain} & 85.2 & 77.9 & 60.4 & 44.0 & 34.0 & 60.3 \\

MIRO$^*$ \cite{cha2022domain} & 85.4 & 78.9 & 69.5 & 45.4 & 44.0 & 64.6 \\
\midrule
Adam + SWAD$^*$ \cite{cha2021swad} & 86.8 & 79.1 & 70.1 & 46.5 & 44.1 & 65.3 \\
AdamW + SWAD  & 87.0 & 78.5 & 70.8 & 46.9 & \textbf{45.0} & 65.6\\
SGD + SWAD & 85.2 & 79.1 & 71.0 & 46.7 & 42.8 & 65.0 \\
FAD (Ours) + SWAD & \textbf{88.5} & \textbf{79.8} & \textbf{71.8} & \textbf{47.5} & \textbf{45.0} & \textbf{66.5} \\
\midrule
Adam + Fishr$^*$ \cite{rame2022fishr} & 85.5 & 78.0 & 68.2 & 46.2 & 44.7 & 64.5 \\
AdamW + Fishr & 85.7 & 77.5 & 68.0 & 46.7 & 43.6 & 64.3 \\
SGD + Fishr & 84.4 & 78.5 & \textbf{69.2} & 46.9 & \textbf{44.4} & 64.7 \\
FAD (Ours) + Fishr & \textbf{88.3} & \textbf{79.6} & \textbf{69.2} & \textbf{48.1} & 43.8 & \textbf{65.8} \\
\midrule
Adam + CORAL$^{*}$ \cite{CORAL16} & 86.0 & 78.9 & 68.7 & 43.7 & 44.5 & 64.5 \\
AdamW + CORAL  & 86.4 & \textbf{79.5} & 69.8 & 45.0 & \textbf{44.9} & 65.1 \\
SGD + CORAL & 85.6 & 78.2 & 69.5 & 45.8 & 44.6  & 64.7\\
FAD (Ours) + CORAL  & \textbf{88.5} & 78.9 & \textbf{70.8} & \textbf{46.1} & \textbf{44.9} & \textbf{65.9} \\
\bottomrule
\end{tabular}}
\label{tab:dg-all}
\end{table*}

\subsection{Computation Overhead}
\label{sec:overhead}
\begin{figure}[t]
    \centering
    \includegraphics[width=\linewidth]{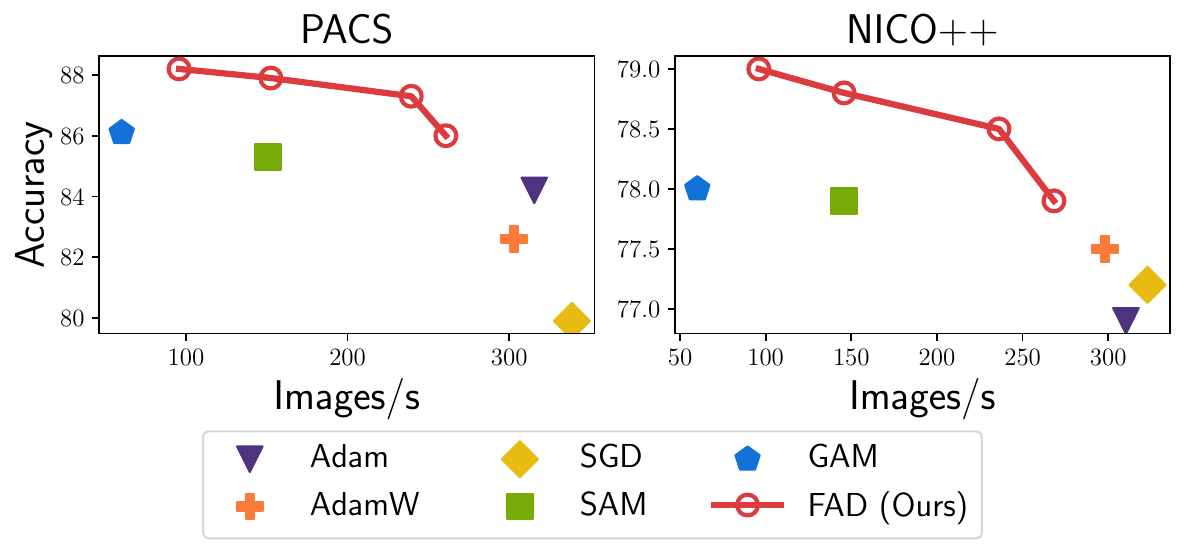}
    \caption{Comparison of computation overhead. Red points on folding lines indicate models trained with FAD for various ratios of iterations (From left to right in each figure, the proportions are 100\%, 50\%, 10\%, 5\%).}
    \label{fig:efficiency}
    \vspace{-15px}
\end{figure}

Since flatness-aware optimization methods require more computation overhead than traditional optimizers, we report the comparison of FAD with other optimizers with more training iterations. We train current optimizers on DG benchmarks for 5,000, 10,000, and 15,000 iterations. We report the performance and required average training time in Table \ref{tab:dg-cost}. Most optimizers achieve noticeable improvements in performance with more training iterations on PACS, VLCS, and OfficeHome. We show that FAD still outperforms current optimizers with fewer training iterations. For example, models trained with FAD for 5,000 iterations on PACS and NICO++ outperform those trained with other optimizers for 15,000 iterations.
This indicates the effectiveness and efficiency of FAD.

Furthermore, as discussed in Section \ref{sect:algorithm-detail}, the FAD gradient can be efficiently calculated via first-order gradient approximation. Yet it still introduces extra computation when calculated in each iteration. Here we investigate optimizing the FAD term only for a subset of iterations in each epoch and optimize the remaining iterations simply with SGD on PACS and NICO++. We change the ratio of the subset with FAD from 5\% to 100\% and report the performance and training speed. We set the batch size to 96 (32 $\times 3$) for PACS and 128 (32 $\times 4$) for NICO++ for all the methods for a fair comparison. As shown in Figure \ref{fig:efficiency}, FAD consistently outperforms SGD, Adam, and AdamW by considerable margins and surpasses SAM and GAM with less training time when setting the ratio of FAD to 10\% and 50\%. Detailed results are in Appendix B.

\begin{table*}[th]
\centering
\caption{The performance and average training time of current optimizers and FAD when training for 5,000, 10,000, 15,000 iterations. \textit{Training Time} is the average time for finishing each dataset with the given iterations.}
\resizebox{0.735\linewidth}{!}{
\begin{tabular}{lc|c|cccc|c}
\toprule
\textbf{Optimizers} & Iterations & Training Time /s & PACS & VLCS & OfficeHome & NICO++ & \textbf{Avg.} \\
\midrule

Adam  & 5,000 & 1613.9 & 84.2 & 77.3 & 67.6 & 76.2 & 76.3 \\
Adam  & 10,000 & 3049.5 & 82.9 & 77.7 & 67.4 & 76.9 & 76.2 \\
Adam  & 15,000 & 4478.6 & 84.0 & 78.4 & 69.6 & 77.3 & 77.3 \\
\midrule
AdamW & 5,000 & 2145.2 & 83.6 & 77.4 & 68.8 & 76.9 & 76.7 \\
AdamW  & 10,000 & 4632.8 & 83.3 & 77.1 & 66.9 & 77.5 & 76.2 \\
AdamW  & 15,000 & 6995.3 & 83.7 & 76.3 & 68.0 & 77.5 & 76.4 \\
\midrule
SGD  & 5,000  & 1864.1 & 79.9 & 78.1 & 68.5 & 76.5 & 75.8 \\
SGD  & 10,000 & 3438.9 &  81.4 & 78.1 & 68.7 & 77.2 & 76.4 \\
SGD  & 15,000 & 4992.0 & 82.6 & 78.7 & 69.6 & 77.0 & 77.0 \\
\midrule
YOGI  & 5,000 & 1388.9 & 81.2 & 77.6 & 68.3 & 77.1 & 76.1 \\
YOGI & 10,000 & 2882.9 & 81.2  & 77.8  & 70.2 & 77.9 & 76.8 \\
YOGI  & 15,000 & 4434.2 & 83.1 & 78.1 & 67.9 & 77.5 & 76.7 \\
\midrule
SAM  & 5,000 & 3325.6 & 85.3 & 78.2 & 68.0 & 77.0 & 77.1 \\
SAM  & 10,000 & 6506.8 & 85.6 & 78.3 & 68.6 & 77.9 & 77.6 \\
SAM  & 15,000 & 9960.2 & 86.1 & 78.9 & 68.8 & 78.1 & 78.0 \\
\midrule
GAM & 5,000 & 8073.6 & 86.1 & 78.5 & 68.2 & 77.4 & 77.6 \\
GAM  & 10,000 & 16210.5 & 86.0 & 79.2 & 68.5 & 78.0 & 78.0 \\
GAM & 15,000 & 23855.1 & 86.5 & 78.6 & 69.0 & 77.8 & 78.0 \\

\midrule
FAD (Ours) & 5,000 & 5057.8 & 88.2 & 78.9 & 69.2 & 78.5 & 78.7 \\
FAD (Ours) & 10,000 & 10450.8 & 88.5 & 79.5 & 69.6 & 79.0 & 79.1 \\
FAD (Ours) & 15,000 &  16255.2 & 88.5 & 79.0 & 69.0 & 79.1 & 78.9 \\
\bottomrule
\end{tabular}}
\label{tab:dg-cost}
\end{table*}

\subsection{The Hessian spectra of selected minima}
\begin{figure*}[t]
    \centering
    \includegraphics[width=\linewidth]{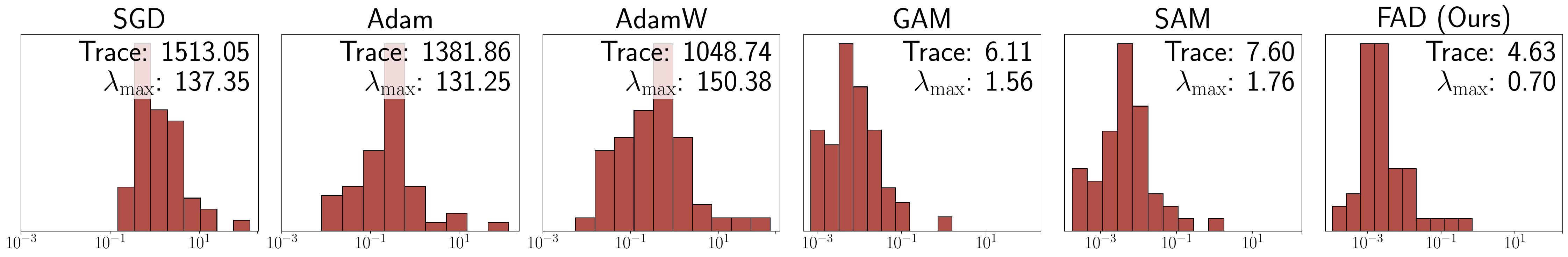}
    \caption{The distribution of top eigenvalues and the trace of Hessian at convergence on PACS with SGD, Adam, AdamW, GAM, SAM, and FAD. \textit{Trace} indicates Hessian trace and \textit{$\lambda_{\max}$} indicates the maximum eigenvalue of Hessian.}
    \label{fig:hessian}
\end{figure*}

Sharpness-aware optimization methods are shown to decrease the curvature of loss landscape \cite{foret2021sharpness,Zhang2023GradientNA}. We investigate how FAD affects the flatness of minima. We consider the maximum eigenvalue of Hessian, which indicates the sharpness across the most critical direction, and the Hessian trace, which measures the expected loss increase under random perturbations to the weights \cite{kaur2022maximum} as the measures of flatness. 

We show the Hessian spectra of ResNet-50 trained on art\_painting, cartoon, and photo domains of PACS for 5,000 iterations with SGD, Adam, AdamW, GAM, SAM, and FAD. Following \cite{Zhang2023GradientNA}, we adopt power iteration \cite{yao2018hessian} to compute the top eigenvalues of Hessian and Hutchinson’s method \cite{bai1996some,avron2011randomized,yao2020pyhessian} to compute the Hessian trace. We report the distribution of the top-50 Hessian eigenvalues and Hessian trace for each method at convergence in Figure~\ref{fig:hessian}. None-flatness-aware optimizers, including SGD, Adam, and AdamW show high Hessian trace and dominant eigenvalues. In contrast, flatness-aware optimizers find minima with significantly lower Hessian spectra. FAD leads to both the lowest Hessian maximum eigenvalue and trace and thus the lowest curvature compared with other optimizers. More results are shown in Appendix B.    
\section{Discussions}
Despite the wide usage in DG benchmarks, Adam is hardly the optimal choice of default optimizer. We show that the best optimizer for different DG methods varies across different datasets. Choosing the right optimizer can help exceed the current benchmarks' limitations. We propose a unified optimization method, flatness-Aware minimization for Domain generalization (FAD), to optimize zeroth-order and first-order flatness efficiently to address the problem. We further show that FAD controls the OOD generalization error and convergences fast. Compared with current optimizers, flatness-aware optimization approaches, including SAM, GAM, and FAD, leads to flatter minima and better generalization. FAD outperforms its counterparts significantly on most DG datasets. 

FAD still has the following limitations which could lead to potential future work. First, the term $\sup_{\bftheta_1, \bftheta_2 \in \caltheta}\left|\calL_{\train}(\bftheta_1, \bftheta_2) - \calL_{\test}(\bftheta_1, \bftheta_2)\right|$ in Equation \eqref{eq:main-bound} indicates the discrepancy between training and test data. Without knowledge of test distribution and extra assumptions, the term cannot be optimized and leads to the optimization of the flatness term. Proper assumptions on the connection between training and test data or the geometry of data may help to learn the connection between flatness and the discrepancy, and thus control the discrepancy for better generalization. Second, We show that Adam is hardly an optimal choice of default optimizer in DG. However, selecting a proper optimizer for each DG method leads to massive overhead. Designing a better DG protocol or learning an indicator for the fitness of optimizers and methods can significantly contribute.  

{\small
\bibliographystyle{ieee_fullname}
\bibliography{egbib}
}

\clearpage
\onecolumn

\appendix
\section{Omitted proofs}
\subsection{Derivation of Equation (3)}
\begin{claim} [Restatement of Equation (3)] \label{claim:eigen-value}
    When a point $\bftheta^*$ is a local minimum of $\errortrain(\bftheta)$ and $\errortrain(\bftheta)$ can be second-order Taylor approximated in the neighbourhood of $\bftheta^*$, then
    \begin{equation} \label{eq:fad-eigen-restate}
      \lambda_{\max}\left(\nabla^2 \errortrain(\bftheta^*)\right) = \frac{\fad_{\rho, \alpha}(\bftheta^*)}{\rho^2\left(1 - \frac{\alpha}{2}\right)}.
    \end{equation}
\end{claim}

\begin{proof} [Derivation of Claim \ref{claim:eigen-value}]
    Suppose a point $\bftheta^*$ is a local minimum of $\errortrain(\bftheta)$ and $\errortrain(\bftheta)$ can be second-order Taylor approximated in the neighbourhood of $\bftheta^*$. Then according to \cite[Lemma 3.3]{zhuang2022surrogate}, we have
    \begin{equation} \label{eq:sam-eigen}
        \lambda_{\max}\left(\nabla^2 \errortrain(\bftheta^*)\right) = \frac{2 \sam_{\rho}(\bftheta^*)}{\rho^2}.
    \end{equation}
    In addition, according to \cite[Lemma 4.1]{Zhang2023GradientNA}, we have
    \begin{equation} \label{eq:gam-eigen}
        \lambda_{\max}\left(\nabla^2 \errortrain(\bftheta^*)\right) = \frac{\gam_{\rho}(\bftheta^*)}{\rho^2}.
    \end{equation}
    Combining Equations \eqref{eq:sam-eigen} and \eqref{eq:gam-eigen}, we have
    \begin{equation}
        \frac{\fad_{\rho, \alpha}(\bftheta^*)}{\rho^2} = \alpha \cdot \frac{\sam(\bftheta^*)}{\rho^2} + (1-\alpha) \cdot \frac{\gam(\bftheta^*)}{\rho^2} = \left(\frac{\alpha}{2} + 1 - \alpha\right)\lambda_{\max}\left(\nabla^2 \errortrain(\bftheta^*)\right) = \left(1 - \frac{\alpha}{2}\right)\lambda_{\max}\left(\nabla^2 \errortrain(\bftheta^*)\right).
    \end{equation}
    Now the claim follows.
\end{proof}

\subsection{Derivation of Equation (9)}
\begin{claim} [Restatement of Equation (9)] \label{claim:gradient}
    The gradient of $\gam_{\rho_t}(\bftheta_t)$ can be approximated as follows
    \begin{equation}
        \begin{aligned}
            \nabla \gam_{\rho_t}(\bftheta_t) & \approx \bfg_{t,3} - \bfg_{t,2}, \quad \text{where} \\
            \tilde{\bftheta}_{t,2} & = \bftheta_t + \rho_t \cdot \frac{\bfg_{t,1} - \bfg_{t,0}}{\|\bfg_{t,1} - \bfg_{t,0}\|}, \quad \bfg_{t,2} = \nabla \errortrain(\tilde{\bftheta}_{t,2}), \\
            \tilde{\bftheta}_{t,3} & = \tilde{\bftheta}_{t,2} + \rho_t \cdot \frac{\bfg_{t,2}}{\|\bfg_{t,2}\|}, \quad \bfg_{t,3} = \nabla \errortrain(\tilde{\bftheta}_{t,3}).
        \end{aligned}
    \end{equation}
    Here
    \begin{equation}
        \bfg_{t,0} = \nabla \errortrain (\bftheta_t), \quad
        \bfg_{t,1} = \nabla \errortrain(\tilde{\bftheta}_{t,1}) \quad \text{where} \quad \tilde{\bftheta}_{t, 1} = \bftheta_t + \rho_t \cdot \frac{\bfg_{t,0}}{\|\bfg_{t,0}\|}.
    \end{equation}
\end{claim}
\begin{proof} [Derivation of Claim \ref{claim:gradient}]
    \cite{Zhang2023GradientNA} propose to approximate the gradient $\nabla \gam_{\rho_t}(\bftheta_t)$ through the following procedure
    \begin{equation*}
        \nabla \gam_{\rho_t}(\bftheta_t) \approx \rho_t \cdot \nabla \left\|\nabla \errortrain(\bftheta^{\text{adv}}_t)\right\|, \bftheta^{\text{adv}}_t = \bftheta_t + \rho_t \cdot \frac{\nabla \left\|\nabla \errortrain(\bftheta_t)\right\|}{\left\|\nabla \|\nabla \errortrain(\bftheta_t)\|\right\|}.
    \end{equation*}
    In addition, $\nabla \|\nabla \errortrain(\bftheta)\|$ is given by the following equation.
    \begin{equation} \label{eq:gam-original}
        \nabla \|\nabla \errortrain(\bftheta)\| = \frac{\nabla^2 \errortrain(\bftheta) \cdot \nabla \errortrain(\bftheta)}{\left\|\errortrain(\bftheta)\right\|}.
    \end{equation}
    However, optimizing the above equations requires the Hessian vector product operation, which is not computation efficient. As a result, inspired by \cite{singh2020woodfisher,zhao2022penalizing},
    one can approximate $\nabla\|\nabla \errortrain(\bftheta)\|$ with first-order gradient as follows.
    \begin{equation} \label{eq:hessian-approx}
        \forall \bftheta \in \caltheta, \quad \nabla\left\|\nabla \errortrain(\bftheta)\right\| \approx \frac{\nabla \errortrain\left(\bftheta + \rho_t \cdot \frac{\nabla \errortrain(\bftheta)}{\|\nabla \errortrain(\bftheta)\|}\right) - \nabla \errortrain(\bftheta)}{\rho_t}.
    \end{equation}
    Applying Equation \eqref{eq:hessian-approx} to the $\bftheta^{\text{adv}}$ term in Equation \eqref{eq:gam-original}, we have
    \begin{equation}
        \bftheta^{\text{adv}} \approx \bftheta_t + \rho_t \cdot \frac{\nabla \errortrain\left(\bftheta_t + \rho_t \cdot \frac{\nabla \errortrain(\bftheta_t)}{\|\nabla \errortrain(\bftheta_t)\|}\right) - \nabla \errortrain(\bftheta_t)}{\left\|\nabla \errortrain\left(\bftheta_t + \rho_t \cdot \frac{\nabla \errortrain(\bftheta_t)}{\|\nabla \errortrain(\bftheta_t)\|}\right) - \nabla \errortrain(\bftheta_t)\right\|} = \bftheta_t + \rho_t \cdot \frac{\bfg_{t,1} - \bfg_{t,0}}{\left\|\bfg_{t,1} - \bfg_{t,0}\right\|}.
    \end{equation}
    We let $\tilde{\bftheta}_{t,2} \triangleq \bftheta^{\text{adv}}$. Now applying Equation \eqref{eq:hessian-approx} to the calculation of $\nabla \|\nabla \errortrain(\tilde{\bftheta}_{t,2})\|$, we can get that
    \begin{equation}
        \begin{aligned}
            \nabla \gam_{\rho}(\bftheta_t) \approx \rho_t \cdot \nabla \|\nabla \errortrain(\tilde{\bftheta}_{t,2})\| & \approx \rho_t \cdot \frac{\nabla \errortrain\left(\tilde{\bftheta}_{t,2}  + \rho_t \cdot \frac{\nabla \errortrain(\tilde{\bftheta}_{t,2} )}{\|\nabla \errortrain(\tilde{\bftheta}_{t,2} )\|}\right) - \nabla \errortrain(\tilde{\bftheta}_{t,2})}{\rho_t} \\
            & = \bfg_{t,3} - \bfg_{t,2},
        \end{aligned}
    \end{equation}
    where
    \begin{equation}
        \bfg_{t,2} = \nabla \errortrain(\tilde{\bftheta}_{t,2}),\quad \bfg_{t,3} = \nabla \errortrain(\tilde{\bftheta}_{t,3}), \quad \text{and} \quad \tilde{\bftheta}_{t,3} = \tilde{\bftheta}_{t,2}  + \rho_t \cdot \frac{\nabla \errortrain(\tilde{\bftheta}_{t,2} )}{\|\nabla \errortrain(\tilde{\bftheta}_{t,2} )\|} = \tilde{\bftheta}_{t,2}  + \rho_t \cdot \frac{\bfg_{t,2}}{\|\bfg_{t,2}\|}.
    \end{equation}
    Now the claim follows.
\end{proof}

\subsection{Proof of Proposition 3.1}
\begin{proof}
    Following the proof of Theorem 1 in \cite{foret2021sharpness}, we can obtain that, by fixing $\sigma = \rho / (\sqrt{d} + \sqrt{\log n})$, we have
    \begin{equation} \label{eq:proof-generalization-main-0}
        \bbE_{\epsilon_i \sim N(0, \sigma^2)}\left[\calE_{\train}(\bftheta+\bfepsilon)\right] \leq \bbE_{\epsilon_i \sim N(0, \sigma^2)}\left[\errortrain(\bftheta+\bfepsilon)\right]+\sqrt{\frac{\frac{1}{4} d \log \left(1+\frac{\|\bftheta\|_2^2}{d \sigma^2}\right)+\frac{1}{4}+\log \frac{n}{\delta}+2 \log (6 n+3 d)}{n-1}}.
    \end{equation}
    Since $\epsilon_i \sim N(0, \sigma^2)$, $\|\bfepsilon\|^2/\sigma^2$ has a chi-square distribution. Therefore, according to Lemma 1 in \cite{laurent2000adaptive}, we have that for any $t > 0$,
    \begin{equation}
        P\left(\|\bfepsilon\|^2/\sigma^2 - d \geq 2 \sqrt{ dt}+2 t\right) \leq \exp (-t).
    \end{equation}
    By fixing $t = \frac{1}{2}\log n$, we can get that with probability at least $1 - 1 / \sqrt{n}$,
    \begin{equation}
      \|\bfepsilon\|^2 \le \sigma^2\left(d + \sqrt{2d \log n} + \log n\right) \le \sigma^2\left(\sqrt{d} + \sqrt{\log n}\right)^2 = \rho^2.
    \end{equation}
    As a result,
    \begin{equation} \label{eq:proof-generalization-main-1}
        \begin{aligned}
            & \, \bbE_{\epsilon_i \sim N(0, \sigma^2)}\left[\errortrain(\bftheta+\bfepsilon)\right] \\
            \le & \, \bbE_{\epsilon_i \sim N(0, \sigma^2)}\left[\errortrain(\bftheta+\bfepsilon) \mid \|\bfepsilon\| \le \rho\right] + \bbE_{\epsilon_i \sim N(0, \sigma^2)}\left[\errortrain(\bftheta+\bfepsilon) \mid \|\bfepsilon\| > \rho\right] \\
            \le & \, \bbE_{\epsilon_i \sim N(0, \sigma^2)}\left[\errortrain(\bftheta+\bfepsilon) \mid \|\bfepsilon\| \le \rho\right] + \frac{M}{\sqrt{n}}. \\
        \end{aligned}
    \end{equation}
    On the one hand, based on the proof of Proposition 4.3 in \cite{Zhang2023GradientNA}, according to the mean value theorem and Cauchy–Schwarz inequality, for any $\bfepsilon$ such that $\|\bfepsilon\| < \rho$, there exists a constant $0 \le c \le 1$, such that
    \begin{equation} \label{eq:proof-generalization-gam}
        \errortrain(\bftheta + \bfepsilon) = \errortrain(\bftheta) + \left(\nabla \errortrain(\bftheta + c \bfepsilon)\right)^\top\bfepsilon \le \errortrain(\bftheta) + \left\|\nabla \errortrain(\bftheta + c \bfepsilon)\right\|\cdot\|\bfepsilon\| \le \errortrain(\bftheta) + \gam_{\rho}(\bftheta).
    \end{equation}
    On the other hand, for any $\bfepsilon$ such that $\|\bfepsilon\| < \rho$, we have
    \begin{equation} \label{eq:proof-generalization-sam}
        \errortrain(\bftheta + \bfepsilon) \le \errortrain(\bftheta) + \sam_{\rho}(\bftheta).
    \end{equation}
    Combining Equations \eqref{eq:proof-generalization-gam} and \eqref{eq:proof-generalization-sam}, we have for any $\bfepsilon$ such that $\|\bfepsilon\| < \rho$ and $0 \le \alpha \le 1$
    \begin{equation} \label{eq:proof-generalization-main-2}
        \errortrain(\bftheta + \epsilon) \le \errortrain(\bftheta) + \alpha \sam_{\rho}(\bftheta) + (1 - \alpha) \gam_{\rho}(\bftheta) = \errortrain(\bftheta) + \fad_{\rho, \alpha}(\bftheta).
    \end{equation}
    Combining Equations \eqref{eq:proof-generalization-main-0}, \eqref{eq:proof-generalization-main-1}, and \eqref{eq:proof-generalization-main-2}, we have
    \begin{equation} \label{eq:proof-generalization-final-1}
        \bbE_{\epsilon_i \sim N(0, \sigma^2)}\left[\calE_{\train}(\bftheta+\bfepsilon)\right] \leq \errortrain(\bftheta) + \fad_{\rho, \alpha}(\bftheta) + \frac{M}{\sqrt{n}} + \sqrt{\frac{\frac{1}{4} d \log \left(1+\frac{\|\bftheta\|_2^2}{d \sigma^2}\right)+\frac{1}{4}+\log \frac{n}{\delta}+2 \log (6 n+3 d)}{n-1}}.
    \end{equation}
    Finally, since we consider the covariate shift scenario, based on Theorem 4.2 in \cite{zhang2022nico++}, we have
    \begin{equation} \label{eq:proof-generalization-final-2}
        \forall \bftheta \in \caltheta, \quad \error_{\test}(\bftheta) \le \error_{\train}(\bftheta) +  \sup_{\bftheta_1, \bftheta_2 \in \caltheta}\left|\calL_{\train}(\bftheta_1, \bftheta_2) - \calL_{\test}(\bftheta_1, \bftheta_2)\right|.
    \end{equation}
    Now the claim follows by combining Equations \eqref{eq:proof-generalization-final-1} and \eqref{eq:proof-generalization-final-2}.
\end{proof}

\subsection{Proof of Theorem 3.2}
We need the following proposition first.

\begin{proposition} \label{prop:convergence-g}
    Suppose the assumptions in Theorem 3.2 hold (with parameters $\gamma, G, M, \eta_0, \rho_0$), then withe learning rate $\eta_t = \eta_0 / \sqrt{t}$ and perturbation radius $\rho_t = \rho_0 / \sqrt{t}$, we have
    \begin{equation}
        \sum_{t=1}^T \bbE\left[\left\|\bfg_{t,0}\right\|^2\right] \le \frac{C_1' + C_2' \log T}{\sqrt{T}}
    \end{equation}
    for some constants $C_1'$ and $C_2'$ that only depend on $\gamma, G, M, \eta_0, \rho_0, \alpha, \beta$.
\end{proposition}

\begin{proof}
    Since $\errortrain(\bftheta)$ is $\gamma$-Lipschitz smooth, we have
    \begin{equation}
        \begin{aligned}
            \errortrain(\bftheta_{t+1}) & \le \errortrain(\bftheta_t) + \left(\nabla \errortrain(\bftheta_t)\right)^{\top} (\bftheta_{t+1} - \bftheta_t) + \frac{\gamma}{2}\left\|\bftheta_t - \bftheta_{t+1}\right\|^2 \\
            & = \errortrain(\bftheta_t) - \eta_t \left(\nabla \errortrain(\bftheta_t)\right)^{\top} \left(\bfg_{t,0} + \beta(\alpha \bfh_{t,0} + (1-\alpha)\bfh_{t,1})\right) + \frac{\gamma\eta_t^2}{2}\left\|\bfg_{t,0} + \beta(\alpha \bfh_{t,0} + (1-\alpha)\bfh_{t,1})\right\|^2.
        \end{aligned}
    \end{equation}
    Now we take the expectation of the above equation conditioned on the events till round $t$. By the assumption, we have $\bbE[\bfg_{t,0}] = \nabla \errortrain(\bftheta_t)$, $\bbE[\bfh_{t,0}] = \bbE[\nabla \errortrain(\tilde{\bftheta}_{t,1})] - \nabla \errortrain(\bftheta_t)$, and $\bbE[\bfh_{t,1}] = \bbE[\nabla \errortrain(\tilde{\bftheta}_{t,3})] - \bbE[\nabla \errortrain(\tilde{\bftheta}_{t,2})]$ we can get that
    \begin{equation} \label{eq:proof-gradient-1}
        \begin{aligned}
            & \, \bbE\left[\errortrain(\bftheta_{t+1})\right] - \errortrain(\bftheta_t) \\
            \le & \,  - \eta_t \left\|\nabla \errortrain(\bftheta_t)\right\|^2 - \eta_t\beta \left(\nabla \errortrain(\bftheta_t)\right)^{\top} \left(\alpha\left(\bbE\left[\nabla \errortrain(\tilde{\bftheta}_{t,1})\right] - \nabla \errortrain(\bftheta_t)\right) + (1-\alpha)\left(\bbE[\nabla \errortrain(\tilde{\bftheta}_{t,3})] - \bbE[\nabla \errortrain(\tilde{\bftheta}_{t,2})]\right)\right) \\
            & \, + \frac{\gamma\eta_t^2}{2}\left\|\bfg_{t,0} + \beta(\alpha \bfh_{t,0} + (1-\alpha)\bfh_{t,1})\right\|^2.
        \end{aligned}
    \end{equation}
    On the one hand, by the Cauchy–Schwarz inequality, the triangle equality of $\|\cdot\|$, and the $\gamma_1$-Lipschitz assumption on $\errortrain(\bftheta)$, we have
    \begin{equation} \label{eq:proof-gradient-2}
        \begin{aligned}
            & - \eta_t\beta\left(\nabla \errortrain(\bftheta_t)\right)^{\top}\left(\alpha\left(\bbE\left[\nabla \errortrain(\tilde{\bftheta}_{t,1})\right] - \nabla \errortrain(\bftheta_t)\right) + (1-\alpha)\left(\bbE[\nabla \errortrain(\tilde{\bftheta}_{t,3})] - \bbE[\nabla \errortrain(\tilde{\bftheta}_{t,2})]\right)\right) \\
            \le & \, \eta_t \beta \left\|\nabla \errortrain(\bftheta_t)\right\|\left\|\alpha\left(\bbE\left[\nabla\errortrain(\tilde{\bftheta}_{t,1})\right] - \nabla\errortrain(\bftheta_t)\right) + (1-\alpha)\left(\bbE[\nabla\errortrain(\tilde{\bftheta}_{t,3})] - \bbE[\nabla\errortrain(\tilde{\bftheta}_{t,2})]\right)\right\| \\
            \le & \, \eta_t \beta G\left(\alpha \left\|\bbE\left[\nabla\errortrain(\tilde{\bftheta}_{t,1}) - \nabla\errortrain(\bftheta_t)\right]\right\| + (1-\alpha)\left\|\bbE[\nabla\errortrain(\tilde{\bftheta}_{t,3}) - \nabla\errortrain(\tilde{\bftheta}_{t,2})]\right\|\right) \\
            \le & \, \eta_t \beta G\left(\alpha \bbE\left[\left\|\nabla\errortrain(\tilde{\bftheta}_{t,1}) - \nabla\errortrain(\bftheta_t)\right\|\right] + (1-\alpha)\bbE\left[\left\|\nabla\errortrain(\tilde{\bftheta}_{t,3}) - \nabla\errortrain(\tilde{\bftheta}_{t,2})\right\|\right]\right) \\
            \le & \, \eta_t \beta G\gamma\left(\alpha \bbE\left[\left\|\tilde{\bftheta}_{t,1} - \bftheta_t\right\|\right] + (1 - \alpha)\bbE\left[\left\|\tilde{\bftheta}_{t,3} - \tilde{\bftheta}_{t,2}\right\|\right]\right) \le \eta_t \rho_t \beta G\gamma.
        \end{aligned}
    \end{equation}
    On the other hand, since $\|\bfg_{t,0}\|, \|\bfg_{t,1}\|, \|\bfg_{t,2}\|, \|\bfg_{t,3}\| \le G$, we have $\|\bfh_{t,0}\|, \|\bfh_{t,1}\| \le 2G$. As a result,
    \begin{equation} \label{eq:proof-gradient-3}
        \begin{aligned}
            & \, \frac{\gamma\eta_t^2}{2}\left\|\bfg_{t,0} + \beta(\alpha \bfh_{t,0} + (1-\alpha)\bfh_{t,1})\right\|^2 \\
            \le & \, \gamma\eta_t^2\left\|\bfg_{t,0}\right\|^2 + \gamma\beta^2 \left\|\alpha \bfh_{t,0} + (1-\alpha)\bfh_{t,1}\right\|^2 \\
            \le & \, \gamma\eta_t^2 \left\|\bfg_{t,0}\right\|^2 + 2\gamma\beta^2\left(\alpha^2\|\bfh_{t,0}\|^2 + (1-\alpha)^2\|\bfh_{t,1}\|^2\right) \\
            \le & \, \gamma\eta_t^2G^2(1 + 8\beta^2(\alpha^2 + (1- \alpha)^2)).
        \end{aligned}
    \end{equation}
    Combining Equations \eqref{eq:proof-gradient-1}, \eqref{eq:proof-gradient-2}, and \eqref{eq:proof-gradient-3}, we can get that
    \begin{equation}
        \eta_t \left\|\bfg_{t,0}\right\|^2 = \eta_t \left\|\nabla \errortrain(\bftheta_t)\right\|^2 \le -\bbE\left[\errortrain(\bftheta_{t+1})\right] + \errortrain(\bftheta_t) + \eta_t\rho_t Z_1 + \eta_t^2 Z_2 
    \end{equation}
    for some constants $Z_1$ and $Z_2$ that depend on $\gamma, G, \alpha, \beta$ only. Now perform telescope sum, take the expectations at each step, and let $\eta_t = \eta_0 / \sqrt{t}$ and $\rho_t = \rho_0 / \sqrt{t}$, we can get that
    \begin{equation}
        \begin{aligned}
            \frac{\eta_0}{\sqrt{T}}\sum_{t=1}^T \left\|\bfg_{t,0}\right\|^2 & \le \eta_t\sum_{t=1}^T \left\|\bfg_{t,0}\right\|^2 \\
            & \le \errortrain\left(\bftheta_1\right) - \bbE\left[\errortrain(\bftheta_{T+1})\right] + Z_1\sum_{t=1}^T \rho_t\eta_t + Z_2 \sum_{t=1}^T \eta_t^2 \\
            & \le \, 2M + Z_1\eta_0\rho_0 \sum_{t=1}^T \frac{1}{t} + Z_2 \eta_0^2\sum_{t=1}^T \frac{1}{t} \\
            & \le \, Z_3 + Z_4 \log T,
        \end{aligned}
    \end{equation}
    for some constants $Z_3$ and $Z_4$ that only depend on $\gamma, G, M, \eta_0, \rho_0, \alpha, \beta$. Now the claim follows.
\end{proof}

Now we can prove Theorem 3.2 with Proposition \ref{prop:convergence-g}.

\begin{proof}[Proof of Theorem 3.2]
    We first observe that
    \begin{equation} \label{eq:proof-theorem-convergence-1}
        \|\bfDelta_t\|^2 = \left\|\bfg_{t,0}+\beta(\alpha \bfh_{t,0} + (1-\alpha)\bfh_{t,1})\right\|^2 \le 2 \|\bfg_{t,0}\|^2 + 2 \beta^2\left\|\alpha \bfh_{t,0} + (1-\alpha)\bfh_{t,1}\right\|^2.
    \end{equation}
    Now we bound the two terms in the right-hand side of the above equation, respectively. The bound on $\sum_{t=1}^T \|\bfg_{t,0}\|^2$ can be obtained from Proposition \ref{prop:convergence-g}. For the second term, at each round $t$, take the expectation conditioned on events till round $t$, we have 
    \begin{equation}
        \begin{aligned}
            & \, \bbE\left[\left\|\alpha \bfh_{t,0} + (1-\alpha)\bfh_{t,1}\right\|^2\right] \\
            \le & \, 2\alpha^2\bbE\left[\left\|\bfh_{t,0}\right\|^2\right] + 2(1-\alpha)^2\bbE\left[\left\|\bfh_{t,1}\right\|^2\right] \\
            \le & \, 2\alpha^2\bbE\left[\left\|\nabla \errortrain(\tilde{\bftheta}_{t,1}) - \nabla \errortrain(\bftheta_t))\right\|^2\right] + 2(1-\alpha)^2\bbE\left[\left\|\nabla \errortrain(\tilde{\bftheta}_{t,3})-\nabla \errortrain(\tilde{\bftheta}_{t,2})\right\|^2\right] \\
            \le & \, 2\alpha^2\gamma^2\bbE\left[\left\|\tilde{\bftheta}_{t,1} - \bftheta_t\right\|^2\right] + 2(1-\alpha)^2\gamma^2\bbE\left[\left\|\tilde{\bftheta}_{t,3}-\tilde{\bftheta}_{t,2}\right\|^2\right] \\
            \le & \, 2\rho_t^2\gamma^2(\alpha^2 + (1-\alpha)^2).
        \end{aligned}
    \end{equation}
    As a result, when $\rho_t = \rho_0 / \sqrt{t}$,
    we have
    \begin{equation} \label{eq:proof-theorem-convergence-2}
        \sum_{t=1}^T \bbE\left[\left\|\alpha \bfh_{t,0} + (1-\alpha)\bfh_{t,1}\right\|^2\right] \le \sum_{t=1}^T 2\rho_t^2\gamma^2(\alpha^2 + (1-\alpha)^2) = 2\rho_0^2\gamma^2(\alpha^2 + (1-\alpha)^2)\sum_{t=1}^T \frac{1}{t} \le Z_1 + Z_2 \log T
    \end{equation}
    for some constants $Z_1$ and $Z_2$ that only depend on $\gamma, \rho_0, \alpha$.

    Now Theorem 3.2 follows by combining Equations \eqref{eq:proof-theorem-convergence-1} and \eqref{eq:proof-theorem-convergence-2} and Proposition \ref{prop:convergence-g}.
\end{proof}
\section{More Experimental Details and Results}

In this section, we report more experimental details and results. 
We present the details of adopted DG datasets, including PACS, VLCS, OfficeHome, TerraInc, DomainNet, and NICO++, for our evaluation, in \ref{sec:datasets}.
We present the training details and hyperparameter search space in \ref{sec:training_details}.
We present the detailed results of current optimizers on DG datasets in \ref{sec:results_of_opts}. We report more results of current optimizers ensembled with DG methods in \ref{sec:ensemble}. We report the detailed results of models trained with current optimizers and FAD for various training iterations in \ref{sec:various_iter}.

\subsection{Datasets}
\label{sec:datasets}

In this subsection, we introduce DG datasets in current benchmarks, including PACS, VLCS, OfficeHome, TerralInc, DomainNet, and NICO++.

\textbf{PACS} \cite{li2017deeper} is a widely used benchmark for domain generalization. It consists of 7 object categories spanning 4 image styles, namely \textsl{photo, art-painting, cartoon}, and \textsl{sketch}. We adopt the protocol in \cite{li2017deeper} to split the training and validation set.

\textbf{VLCS} \cite{fang2013unbiased} consists of 5 object categories shared by the PASCAL VOC 2007, LabelMe, Caltech and Sun datasets. We follow the standard protocol of \cite{ghifary2015domain} and divide each domain into a training set (70\%) and validation set (30\%) randomly. 

\textbf{OfficeHome} \cite{venkateswara2017deep} includes 65 categories and 4 domains, namely \textsl{art, clipart, product}, and \textsl{real}. It contains 15,588 samples.

\textbf{Terra Incognita} \cite{beery2018recognition} contains photographs of wild animals shot at 4 locations, namely L100, L38, L43, and L46. We adopt the subset of it used in DomainBed\cite{gulrajani2020search} and it consists of 10 categories and 24,788 samples.

\textbf{DomainNet} \cite{peng2019moment} contains 6 domains, including \textsl{clipart, infograph, painting, quickdraw, real}, and \textsl{sketch}. It contains 345 categories and 586,575 samples. 

\textbf{NICO++} \cite{zhang2022nico++} is a large scale OOD dataset. It contains more than 200,000 natural images. It consists of 80 categories and 10 common contexts for DG tasks.

\subsection{Training details}
\label{sec:training_details}
As introduced in Section Experiments, we follow the basic training and evaluation protocol introduced in \cite{gulrajani2020search}, where the information of test data is unavailable for hyperparameter search. 
We train all the models on DomainNet for 15,000 iterations as suggested in \cite{cha2021swad}, 10,000 iterations on NICO++, and 5,000 iterations on other datasets unless otherwise noted. 
For datasets except for NICO++, we follow the leave-one-out protocol in \cite{gulrajani2020search} where one domain is chosen as the target domain and the remaining domains as the training domain.
For NICO++, we choose two domains as target domains in each run and train models on the remaining four domains. Following the official combination \cite{zhang2022nico++}, we select \{\textit{autumn
}, \textit{rock}\}, \{\textit{dim},\textit{grass}\}, \{\textit{outdoor}, \textit{water}\} as the target domain pairs. Detailed training and test splits of NICO++ are shown in Table \ref{tab:nico_split}. We use the official training subset of each selected training domain for training and test subset for test. We report the results of all the domains together in Table \ref{tab:nico10k}.

\begin{table*}[th]
\centering
\caption{The detailed split of training and test domains of NICO++.}
\begin{tabular}{l|c}
\toprule
Training domains & Test domains \\
\midrule
  dim, grass, outdoor, water & autumn, rock \\ 
 autumn, rock, outdoor, water & dim, grass \\
autumn, rock, dim, grass &  outdoor, water \\

\bottomrule
\end{tabular}
\label{tab:nico_split}
\end{table*}

All the models are evaluated following the protocol in DomainBed. The search space of hyperparameters is shown in Table \ref{tab:hyper_space}.

\begin{table*}[th]
\centering
\caption{The search space of hyperparameters.}
\begin{tabular}{l|cc}
\toprule
Parameter & Default value & Search Distribution \\
\midrule
batch size & 32 & 2$^{\rm Uniform(3, 5.5)}$ \\
learning rate & 0.00005 & 10$^{\rm Uniform(-5, -3.5)}$ \\
momentum & 0.9 & 10$^{\rm Uniform(-1, 0)}$ \\
weight decay & 0.0001 & 10$^{\rm Uniform(-6, -3)}$ \\
\bottomrule
\end{tabular}
\label{tab:hyper_space}
\end{table*}

For the search space of algorithm-specific hyperparameters of SAM \cite{du2021efficient} and GAM \cite{Zhang2023GradientNA}, we follow the settings in their original papers, where $\rho$ is searched in \{0.01, 0.05, 0.1, 0.2, 0.5, 1.0\} and $\alpha$ in \{0.1, 0.2, 0.5, 1.0, 2.0, 3.0, ..., 10.0\}. There are three unique hyperparameters in GAM, namely $\rho$, $\alpha$, and $\beta$. We search $\rho$ in \{0.05, 0.1, 0.2, 0.5, 1.0, 2.0\}, $\alpha$ in \{0.1, 0.2, ..., 1.0\}, and $\beta$ in \{0.01, 0.05, 0.1, 0.2, 0.5, 1.0\}. Please note that all the hyperparameters are randomly selected following the protocol in DomainBed and no information of test data is known in the training and model selection phase. For SAM, GAM, and FAD, we use SGD, which is the naive optimizer compared with adaptive or second-order aware optimizers, and the base optimizer.      

\subsection{Detailed results of current optimizers on DG datasets with the evaluation protocol in DomainBed.}
\label{sec:results_of_opts}
We report the test accuracy of current optimizers and FAD on different target domains of DG datasets.  

\subsubsection{PACS}
We report the detailed results on PACS in Table \ref{tab:pacs5k}. FAD consistently outperforms its counterparts on all domains.
\begin{table*}[th]
\centering
\caption{The comparison of optimizers on PACS. All the models are trained for 5,000 iterations and evaluated following the protocol in DomainBed~\cite{gulrajani2020search}. The best results for each domain are highlighted in bold font.}
\begin{tabular}{lcccc|c}
\toprule
\textbf{Algorithm} & art & cartoon & photo & sketch  & \textbf{Avg.} \\
\midrule

Adam  & 88.0 \footnotesize{$\pm 1.2$} &  79.7 \footnotesize{$\pm 0.5$} & 96.7 \footnotesize{$\pm 0.4$} & 72.7 \footnotesize{$\pm 0.9$} & 84.3 \\
AdamW  & 84.1 \footnotesize{$\pm 1.5$} & 80.7 \footnotesize{$\pm 1.2$}  & 96.9 \footnotesize{$\pm 0.4$} & 72.8 \footnotesize{$\pm 0.6$} & 83.6 \\
SGD  & 85.1 \footnotesize{$\pm 0.4$} & 76.0 \footnotesize{$\pm 0.3$}  & 98.3 \footnotesize{$\pm 0.4$} & 60.3 \footnotesize{$\pm 6.1$} & 79.9 \\
YOGI  & 84.4 \footnotesize{$\pm 1.7$} & 79.7 \footnotesize{$\pm 0.6$}  & 95.8 \footnotesize{$\pm 0.3$} & 65.1 \footnotesize{$\pm 1.5$} & 81.2 \\
AdaBelief  & 85.4 \footnotesize{$\pm 2.2$} & 80.4 \footnotesize{$\pm 1.1$}  & 97.4 \footnotesize{$\pm 0.7$} & 75.1 \footnotesize{$\pm 1.4$} & 84.6\\
AdaHessian  & 88.4 \footnotesize{$\pm 0.6$} &  80.0 \footnotesize{$\pm 0.9$} & 97.7 \footnotesize{$\pm 0.4$} & 71.7 \footnotesize{$\pm 4.1$} & 84.5 \\
SAM  & 85.7 \footnotesize{$\pm 1.2$} & 81.0 \footnotesize{$\pm 1.4$}  & 97.1 \footnotesize{$\pm 0.2$} & 77.4 \footnotesize{$\pm 1.8$} & 85.3 \\
GAM  & 85.9 \footnotesize{$\pm 0.9$} & 81.3 \footnotesize{$\pm 1.6$}  & 98.2 \footnotesize{$\pm 0.4$} & 79.0 \footnotesize{$\pm 2.1$} & 86.1 \\
\midrule
FAD (ours)  & \textbf{88.5} \footnotesize{$\pm 0.5$} & \textbf{83.0} \footnotesize{$\pm 0.8$}  & \textbf{98.4} \footnotesize{$\pm 0.2$} & \textbf{82.8} \footnotesize{$\pm 0.9$} & \textbf{88.2} \\

\bottomrule
\end{tabular}
\label{tab:pacs5k}
\end{table*}

\subsubsection{VLCS}
We report the detailed results on VLCS in Table \ref{tab:vlcs5k}. FAD outperforms its counterparts on 3 out of 4 domains and achieves the highest average accuracy.

\begin{table*}[th]
\centering
\caption{The comparison of optimizers on VLCS. All the models are trained for 5,000 iterations and evaluated following the protocol in DomainBed~\cite{gulrajani2020search}. The best results for each domain are highlighted in bold font.}
\begin{tabular}{lcccc|c}
\toprule
\textbf{Algorithm} & Caltech & LabelMe & SUN & VOC  & \textbf{Avg.} \\
\midrule

Adam & 98.9 \footnotesize{$\pm 0.4$} &  65.9 \footnotesize{$\pm 1.5$} & 71.0 \footnotesize{$\pm 1.6$} & 74.5 \footnotesize{$\pm 2.0$} & 77.3 \\
AdamW  & 98.3 \footnotesize{$\pm 0.1$} &  65.1 \footnotesize{$\pm 1.7$} & 70.9 \footnotesize{$\pm 1.3$} & 75.2 \footnotesize{$\pm 1.5$} & 77.4 \\
SGD  & 98.4 \footnotesize{$\pm 0.2$} &  64.7 \footnotesize{$\pm 0.7$} & 72.5 \footnotesize{$\pm 0.8$} & 76.6 \footnotesize{$\pm 0.8$} & 78.1 \\
YOGI  & 98.1 \footnotesize{$\pm 0.7$} &  63.9 \footnotesize{$\pm 1.2$} & 72.5 \footnotesize{$\pm 1.6$} & 75.7 \footnotesize{$\pm 1.2$} & 77.6 \\
AdaBelief  & 98.0 \footnotesize{$\pm 0.1$} &  63.9 \footnotesize{$\pm 0.4$} & 73.4 \footnotesize{$\pm 1.0$} & \textbf{78.2} \footnotesize{$\pm 1.8$} & 78.4 \\
AdaHessian  & \textbf{99.1} \footnotesize{$\pm 0.3$} &  65.0 \footnotesize{$\pm 1.7$} & 72.7 \footnotesize{$\pm 1.3$} & 77.7 \footnotesize{$\pm 1.0$} & 78.6 \\
SAM  & 98.5 \footnotesize{$\pm 1.0$} & 66.2 \footnotesize{$\pm 1.6$}  & 72.0 \footnotesize{$\pm 1.0$} & 76.1 \footnotesize{$\pm 1.0$} &  78.2 \\
GAM  & 98.8 \footnotesize{$\pm 0.6$} & 65.1 \footnotesize{$\pm 1.2$}  & 72.9 \footnotesize{$\pm 1.0$} & 77.2 \footnotesize{$\pm 1.9$} & 78.5 \\
\midrule
FAD (ours)  & \textbf{99.1} \footnotesize{$\pm 0.5$} & \textbf{66.8} \footnotesize{$\pm 0.9 $}  & \textbf{73.6} \footnotesize{$\pm 1.0 $} & 76.1 \footnotesize{$\pm 1.3$} &  \textbf{78.9} \\

\bottomrule
\end{tabular}
\label{tab:vlcs5k}
\end{table*}

\subsubsection{OfficeHome}
We report the detailed results on OfficeHome in Table \ref{tab:office5k}. FAD outperforms its counterparts on 3 out of 4 domains and achieves the highest average accuracy.
\begin{table*}[th]
\centering
\caption{The comparison of optimizers on OfficeHome. All the models are trained for 5,000 iterations and evaluated following the protocol in DomainBed~\cite{gulrajani2020search}. The best results for each domain are highlighted in bold font.}
\begin{tabular}{lcccc|c}
\toprule
\textbf{Algorithm} & Art & Clipart & Product & Real-World  & \textbf{Avg.} \\
\midrule

Adam  & 63.9 \footnotesize{$\pm 0.8$} & 48.1 \footnotesize{$\pm 0.6$} & 77.0 \footnotesize{$\pm 0.9$} & 81.8 \footnotesize{$\pm 1.6$} & 67.6 \\
AdamW  & \textbf{66.1} \footnotesize{$\pm 0.7$} & 48.7 \footnotesize{$\pm 0.6$} & 76.6 \footnotesize{$\pm 0.8$} & 83.6 \footnotesize{$\pm 0.4$} & 68.8 \\
SGD  & 65.3 \footnotesize{$\pm 0.8$} & 48.8 \footnotesize{$\pm 1.4$} & 76.7 \footnotesize{$\pm 0.3$} & 83.0 \footnotesize{$\pm 0.7$} & 68.5 \\
YOGI  & 63.5 \footnotesize{$\pm 1.0$} & 49.2 \footnotesize{$\pm 1.2$} & 76.2 \footnotesize{$\pm 0.5$} & 84.5 \footnotesize{$\pm 0.6$} & 68.3 \\
AdaBelief  & 65.6 \footnotesize{$\pm 2.0$} & 48.1 \footnotesize{$\pm 0.9$} & 74.8 \footnotesize{$\pm 0.8$} & 83.6 \footnotesize{$\pm 0.9$} & 68.0\\
AdaHessian  & 63.0 \footnotesize{$\pm 2.9$} & 50.0 \footnotesize{$\pm 1.4$} & 77.7 \footnotesize{$\pm 0.8$} & 83.0 \footnotesize{$\pm 0.5$} & 68.4 \\
SAM  & 63.5 \footnotesize{$\pm 1.2$} & 48.6 \footnotesize{$\pm 0.9$}  & 77.0 \footnotesize{$\pm 0.8$} & 82.9 \footnotesize{$\pm 1.3$} & 68.0 \\
GAM  & 63.0 \footnotesize{$\pm 1.2$} & 49.8 \footnotesize{$\pm 0.5$}  & 77.6 \footnotesize{$\pm 0.6$}  & 82.4 \footnotesize{$\pm 1.0$} & 68.2 \\
\midrule
FAD (ours)  &  63.5 \footnotesize{$\pm 1.0$} & \textbf{50.3} \footnotesize{$\pm 0.8$}  & \textbf{78.0} \footnotesize{$\pm 0.4$} & \textbf{85.0} \footnotesize{$\pm 0.6$} &  \textbf{69.2}\\

\bottomrule
\end{tabular}
\label{tab:office5k}
\end{table*}

\subsubsection{TerraInc}
We report the detailed results on TerraInc in Table \ref{tab:terra5k}. FAD outperforms its counterparts on 3 out of 4 domains and achieves the highest average accuracy.
\begin{table*}[th]
\centering
\caption{The comparison of optimizers on TerraInc. All the models are trained for 5,000 iterations and evaluated following the protocol in DomainBed~\cite{gulrajani2020search}. The best results for each domain are highlighted in bold font.}
\begin{tabular}{lcccc|c}
\toprule
\textbf{Algorithm} & L100 & L38 & L43 & L46  & \textbf{Avg.} \\
\midrule

Adam  & 42.2 \footnotesize{$\pm 3.4$} & 40.7 \footnotesize{$\pm 1.2$} & 59.9 \footnotesize{$\pm 0.2$} & 35.0 \footnotesize{$\pm 2.8$} & 44.4  \\
AdamW  & 44.2 \footnotesize{$\pm 6.8$} & 39.8 \footnotesize{$\pm 1.9$} & 60.3 \footnotesize{$\pm 2.0$} & 36.6 \footnotesize{$\pm 1.8$} & 45.2 \\
SGD  & 41.8 \footnotesize{$\pm 5.8$} & 39.8 \footnotesize{$\pm 3.9$} & 60.5 \footnotesize{$\pm 2.2$} & \textbf{37.5} \footnotesize{$\pm 1.1$} & 44.9 \\
YOGI  & 43.9 \footnotesize{$\pm 2.2$} & 42.5 \footnotesize{$\pm 2.6$} & 60.5 \footnotesize{$\pm 1.1$} & 34.8 \footnotesize{$\pm 1.6$} & 45.4 \\
AdaBelief  & 42.6 \footnotesize{$\pm 6.7$} & 43.0 \footnotesize{$\pm 2.0$} & 60.2 \footnotesize{$\pm 1.3$} & 35.1 \footnotesize{$\pm 0.3$} & 45.2 \\
AdaHessian  & 42.5 \footnotesize{$\pm 4.8$} & 39.5 \footnotesize{$\pm 1.0$} & 58.4 \footnotesize{$\pm 2.6$} & 37.3 \footnotesize{$\pm 0.8$} & 44.4 \\
SAM  & 42.9 \footnotesize{$\pm 3.5$} &  43.0 \footnotesize{$\pm 2.2$} & 60.5 \footnotesize{$\pm 1.6$} & 36.4 \footnotesize{$\pm 1.2$} & 45.7\\
GAM  & 42.2 \footnotesize{$\pm 2.6$} & 42.9 \footnotesize{$\pm 1.7$}  & 60.2 \footnotesize{$\pm 1.8$} & 35.5 \footnotesize{$\pm 0.7$} & 45.2\\
\midrule
FAD (ours)  & \textbf{44.3} \footnotesize{$\pm 2.2$} & \textbf{43.5} \footnotesize{$\pm 1.7$}  & \textbf{60.9} \footnotesize{$\pm 2.0$} & 34.1 \footnotesize{$\pm 0.5$} &  \textbf{45.7}\\

\bottomrule
\end{tabular}
\label{tab:terra5k}
\end{table*}

\subsubsection{DomainNet}
We report the detailed results on DomainNet in Table \ref{tab:domainnet15k}. FAD outperforms its counterparts on 4 out of 6 domains and achieves the highest average accuracy.
\begin{table*}[th]
\centering
\caption{The comparison of optimizers on DomainNet. All the models are trained for 15,000 iterations and evaluated following the protocol in DomainBed~\cite{gulrajani2020search}. The best results for each domain are highlighted in bold font.}
\begin{tabular}{lcccccc|c}
\toprule
\textbf{Algorithm} & clip & info & paint & quick  & real & sketch&  \textbf{Avg.} \\
\midrule

Adam  & 63.0 \footnotesize{$\pm 0.3$} &  20.2 \footnotesize{$\pm 0.4$} & 49.1 \footnotesize{$\pm 0.1$} & 13.0 \footnotesize{$\pm 0.3$} & 62.0 \footnotesize{$\pm 0.4$} & 50.7 \footnotesize{$\pm 0.1$} & 43.0 \\
AdamW  & 63.0 \footnotesize{$\pm 0.6$} &  20.6 \footnotesize{$\pm 0.2$} & 49.6 \footnotesize{$\pm 0.0$} & 13.0 \footnotesize{$\pm 0.2$} & 63.6 \footnotesize{$\pm 0.2$} & 50.4 \footnotesize{$\pm 0.1$} & 43.4  \\
SGD  & 61.3 \footnotesize{$\pm 0.2$} &  20.4 \footnotesize{$\pm 0.2$} & 49.4 \footnotesize{$\pm 0.2$} & 12.6 \footnotesize{$\pm 0.1$} & \textbf{65.7} \footnotesize{$\pm 0.0$} & 49.6 \footnotesize{$\pm 0.2$} & 43.2 \\
YOGI  & 63.3 \footnotesize{$\pm 0.1$} &  20.6 \footnotesize{$\pm 0.1$} & 50.1 \footnotesize{$\pm 0.3$} & 13.2 \footnotesize{$\pm 0.3$} & 62.8 \footnotesize{$\pm 0.1$} & 51.0 \footnotesize{$\pm 0.2$} & 43.5 \\
AdaBelief  & 63.5 \footnotesize{$\pm 0.2$} &  20.5 \footnotesize{$\pm 0.1$} & 50.0 \footnotesize{$\pm 0.3$} & 13.2 \footnotesize{$\pm 0.3$} & 63.1 \footnotesize{$\pm 0.1$} & 50.7 \footnotesize{$\pm 0.1$} & 43.5 \\
AdaHessian  & 63.3 \footnotesize{$\pm 0.2$} &  21.4 \footnotesize{$\pm 0.1$} & \textbf{50.8} \footnotesize{$\pm 0.3$} & 13.6 \footnotesize{$\pm 0.1$} & \textbf{65.7} \footnotesize{$\pm 0.1$} & 51.4 \footnotesize{$\pm 0.2$} & 44.4 \\
SAM  & 63.3 \footnotesize{$\pm 0.1$} & 20.3 \footnotesize{$\pm 0.3$} & 50.0 \footnotesize{$\pm 0.3$} & 13.6 \footnotesize{$\pm 0.2$} & 63.6 \footnotesize{$\pm 0.3$} & 49.6 \footnotesize{$\pm 0.4$} & 43.4 \\
GAM  & 63.0 \footnotesize{$\pm 0.5$} & 20.2 \footnotesize{$\pm 0.2$} & 50.3 \footnotesize{$\pm 0.1$} & 13.2 \footnotesize{$\pm 0.3$} & 64.5 \footnotesize{$\pm 0.2$} & 51.6 \footnotesize{$\pm 0.5$} & 43.8 \\
\midrule
FAD (ours) & \textbf{64.1} \footnotesize{$\pm 0.3$} & \textbf{21.9} \footnotesize{$\pm 0.2$} & 50.6 \footnotesize{$\pm 0.3$} & \textbf{14.2} \footnotesize{$\pm 0.4$} & 63.6 \footnotesize{$\pm 0.1$} & \textbf{52.2} \footnotesize{$\pm 0.2$} & \textbf{44.4} \\

\bottomrule
\end{tabular}
\label{tab:domainnet15k}
\end{table*}

\subsubsection{NICO++}
We report the detailed results on NICO++ in Table \ref{tab:nico10k}. FAD consistently outperforms its counterparts on all domains.
\begin{table*}[th]

\centering
\caption{The comparison of optimizers on NICO++. All the models are trained for 10,000 iterations and evaluated following the protocol in DomainBed~\cite{gulrajani2020search}. The best results for each domain are highlighted in bold font.}
\begin{tabular}{lcccccc|c}
\toprule
\textbf{Algorithm} & Autumn & Rock & Dim & Grass  & Outdoor & Water &\textbf{Avg.} \\
\midrule

Adam  & 81.9 \footnotesize{$\pm 0.3$} &  79.8 \footnotesize{$\pm 0.2$} & 72.4 \footnotesize{$\pm 0.3$} & 82.3 \footnotesize{$\pm 0.2$} & 76.8 \footnotesize{$\pm 0.5$} & 71.0 \footnotesize{$\pm 0.3$} & 76.9  \\
AdamW  & 81.6 \footnotesize{$\pm 0.2$} &  79.5 \footnotesize{$\pm 0.2$} & 72.4 \footnotesize{$\pm 0.2$} & 82.7 \footnotesize{$\pm 0.3$} & 77.4 \footnotesize{$\pm 0.3$} & 71.4 \footnotesize{$\pm 0.6$} & 77.5 \\
SGD  & 81.5 \footnotesize{$\pm 0.5$} &  79.3 \footnotesize{$\pm 0.4$} & 71.7 \footnotesize{$\pm 0.2$} & 82.5 \footnotesize{$\pm 0.1$} & 77.0 \footnotesize{$\pm 0.4$} & 71.3 \footnotesize{$\pm 0.6$} & 77.2 \\
YOGI  & 82.1 \footnotesize{$\pm 0.3$} &  80.1 \footnotesize{$\pm 0.3$} & 72.8 \footnotesize{$\pm 0.4$} & 83.0 \footnotesize{$\pm 0.1$} & 77.4 \footnotesize{$\pm 0.2$} & 71.8 \footnotesize{$\pm 0.3$} & 77.9 \\
AdaBelief  & 81.4 \footnotesize{$\pm 0.2$} &  79.1 \footnotesize{$\pm 0.2$} & 72.2 \footnotesize{$\pm 0.3$} & 82.8 \footnotesize{$\pm 0.3$} & 77.5 \footnotesize{$\pm 0.1$} & 71.4 \footnotesize{$\pm 0.3$} & 77.4 \\
AdaHessian  & 82.0 \footnotesize{$\pm 0.4$} &  80.1 \footnotesize{$\pm 0.3$} & 73.0 \footnotesize{$\pm 0.0$} & 82.5 \footnotesize{$\pm 0.3$} & 77.5 \footnotesize{$\pm 0.3$} & 71.2 \footnotesize{$\pm 0.4$} & 77.7 \\
SAM  & 82.6 \footnotesize{$\pm 0.3$} & 80.6 \footnotesize{$\pm 0.1$} & 72.3 \footnotesize{$\pm 0.4$} & 82.9 \footnotesize{$\pm 0.3$} & 77.0 \footnotesize{$\pm 0.1$} & 72.0 \footnotesize{$\pm 0.4$} & 77.9 \\
GAM  & 82.3 \footnotesize{$\pm 0.1$} & 80.8 \footnotesize{$\pm 0.2$} & 72.5 \footnotesize{$\pm 0.2$} & 82.9 \footnotesize{$\pm 0.4$} & 77.4 \footnotesize{$\pm 0.3$} & 72.1 \footnotesize{$\pm 0.5$} & 78.0 \\
\midrule
FAD (ours) & \textbf{83.5} \footnotesize{$\pm 0.3$} & \textbf{81.8} \footnotesize{$\pm 0.4$} & \textbf{74.2} \footnotesize{$\pm 0.2$} & \textbf{83.3} \footnotesize{$\pm 0.3$} & \textbf{78.0} \footnotesize{$\pm 0.1$} & \textbf{73.2} \footnotesize{$\pm 0.5$} & \textbf{79.0} \\

\bottomrule
\end{tabular}
\label{tab:nico10k}
\end{table*}

\subsection{Ensemble with DG methods.}
\label{sec:ensemble}
We present more results of optimizers ensembled with DG methods in Table \ref{tab:ensemble}. The results are aligned with them in the main paper. Different optimizers seem to favor different DG methods and FAD consistently outperforms other optimizers with all the methods across various datasets.

\begin{table*}[th]
\centering
\caption{Ensemble with domain generalization methods. Numbers for methods marked with $^*$ and all the combinations of DG methods with optimizers other than Adam are reproduced results. Other results are from the original literature and DomainBed (donated with $^{\dagger}$).}
\resizebox{0.7\linewidth}{!}{
\begin{tabular}{lccccc|c}
\toprule
\textbf{Algorithm} & PACS & VLCS & OfficeHome & TerraInc & DomainNet & \textbf{Avg.} \\
\midrule
MASF \cite{NEURIPS2019_2974788b} & 82.7 & - & - & - & - & - \\
DMC \cite{Chattopadhyay20} & 83.4 & - & - & - & 43.6 & - \\
MetaReg \cite{NEURIPS2018_647bba34} & 83.6 & - & - & - & 43.6 & - \\
ER \cite{NEURIPS2020_b98249b3} & 85.3 & - & - & - & - & - \\
pAdalN \cite{AdaIN21} & 85.4 & - & - & - & - & - \\
EISNet \cite{Wang2020LearningFE} & 85.8 & - & - & - & - & - \\
DSON \cite{Seonguk20} & 86.6 & - & - & - & - & - \\
ERM$^{\dagger}$ \cite{vapnik1999overview} & 85.5 & 77.5 & 66.5 & 46.1 & 40.9 & 63.3 \\
ERM$^*$ (with Adam) & 84.2 & 77.3 & 67.6 & 44.4 & 43.0 & 63.3 \\
IRM$^{\dagger}$ \cite{Arjovsky19} & 83.5 & 78.6 & 64.3 & 47.6 & 33.9 & 61.6 \\
GroupDRO$^{\dagger}$ \cite{Sagawa20Distributionally} & 84.4 & 76.7 & 66.0 & 43.2 & 33.3 & 60.7 \\
I-Mixup$^{\dagger}$ \cite{XuMinghao19,YanShen20,WangYufei20} & 84.6 & 77.4 & 68.1 & 47.9 & 39.2 & 63.4 \\
MLDG$^{\dagger}$ \cite{LiDa17} & 84.9 & 77.2 & 66.8 & 47.8 & 41.2 & 63.6 \\
MMD$^{\dagger}$ \cite{LiHaoliang18} & 84.7 & 77.5 & 66.4 & 42.2 & 23.4 & 58.8 \\
DANN$^{\dagger}$ \cite{GaninYaroslav15} & 83.7 & 78.6 & 65.9 & 46.7 & 38.3 & 62.6 \\
CDANN$^{\dagger}$ \cite{LiYa18} & 82.6 & 77.5 & 65.7 & 45.8 & 38.3 & 62.0 \\
MTL$^{\dagger}$ \cite{BlanchardGilles17} & 84.6 & 77.2 & 66.4 & 45.6 & 40.6 & 62.9 \\
SagNet$^{\dagger}$ \cite{Nam2021ReducingDG} & 86.3 & 77.8 & 68.1 & 48.6 & 40.3 & 64.2 \\
ARM$^{\dagger}$ \cite{zhang2021adaptive} & 85.1 & 77.6 & 64.8 & 45.5 & 35.5 & 61.7 \\
VREx$^{\dagger}$ \cite{KruegerDavid20} & 84.9 & 78.3 & 66.4 & 46.4 & 33.6 & 61.9 \\
RSC$^{\dagger}$ \cite{Huangzeyi20} & 85.2 & 77.1 & 65.5 & 46.6 & 38.9 & 62.7 \\
Mixstyle \cite{zhou2021domain} & 85.2 & 77.9 & 60.4 & 44.0 & 34.0 & 60.3 \\

MIRO$^*$ \cite{cha2022domain} & 85.4 & 78.9 & 69.5 & 45.4 & 44.0 & 64.6 \\
\midrule
Adam + SWAD$^*$ \cite{cha2021swad} & 86.8 & 79.1 & 70.1 & 46.5 & 44.1 & 65.3 \\
AdamW + SWAD  & 87.0 & 78.5 & 70.8 & 46.9 & \textbf{45.0} & 65.6\\
SGD + SWAD & 85.2 & 79.1 & 71.0 & 46.7 & 42.8 & 65.0 \\
FAD (Ours) + SWAD & \textbf{88.5} & \textbf{79.8} & \textbf{71.8} & \textbf{47.5} & \textbf{45.0} & \textbf{66.5} \\
\midrule
Adam + Fishr$^*$ \cite{rame2022fishr} & 85.5 & 78.0 & 68.2 & 46.2 & 44.7 & 64.5 \\
AdamW + Fishr & 85.7 & 77.5 & 68.0 & 46.7 & 43.6 & 64.3 \\
SGD + Fishr & 84.4 & 78.5 & \textbf{69.2} & 46.9 & \textbf{44.4} & 64.7 \\
FAD (Ours) + Fishr & \textbf{88.3} & \textbf{79.6} & \textbf{69.2} & \textbf{48.1} & 43.8 & \textbf{65.8} \\
\midrule
Adam + CORAL$^{*}$ \cite{CORAL16} & 86.0 & 78.9 & 68.7 & 43.7 & 44.5 & 64.5 \\
AdamW + CORAL  & 86.4 & \textbf{79.5} & 69.8 & 45.0 & \textbf{44.9} & 65.1 \\
SGD + CORAL & 85.6 & 78.2 & 69.5 & 45.8 & 44.6  & 64.7\\
FAD (Ours) + CORAL  & \textbf{88.5} & 78.9 & \textbf{70.8} & \textbf{46.1} & \textbf{44.9} & \textbf{65.9} \\
\midrule
Adam + EoA$^*$ \cite{arpit2021ensemble} & 87.0 & 79.4 & 70.2 & 46.2 & 44.5 & 65.5 \\
AdamW + EoA  & 87.5 & 78.4 & 71.9 & 47.7 & 45.2 & 66.1\\
SGD + EoA & 86.0 & 79.2 & 71.5 & 46.8 & 42.5 & 65.2 \\
FAD (Ours) + EoA & \textbf{88.9} & \textbf{79.7} & \textbf{72.0} & \textbf{47.6} & \textbf{45.3} & \textbf{66.7} \\
\midrule
Adam + RSC$^{*}$ \cite{huang2020self} & 84.5 & 77.9 & 65.7 & 44.5 & 42.8 & 63.1 \\
AdamW + RSC  & 83.4 & 77.5 & 66.3 & 45.1 & 42.4 & 62.9 \\
SGD + RSC & 82.6 & \textbf{78.1} & 67.0 & 43.9 & 43.5  & 63.0\\
FAD (Ours) + RSC  & \textbf{86.9} & 77.6 & \textbf{68.6} & \textbf{46.2} & \textbf{44.1} & \textbf{64.7} \\

\bottomrule
\end{tabular}}
\label{tab:dg-all}
\label{tab:ensemble}
\end{table*}

\subsection{Optimizers with various training iterations}
\label{sec:various_iter}
\begin{figure}[t]
    \centering
    \includegraphics[width=\linewidth]{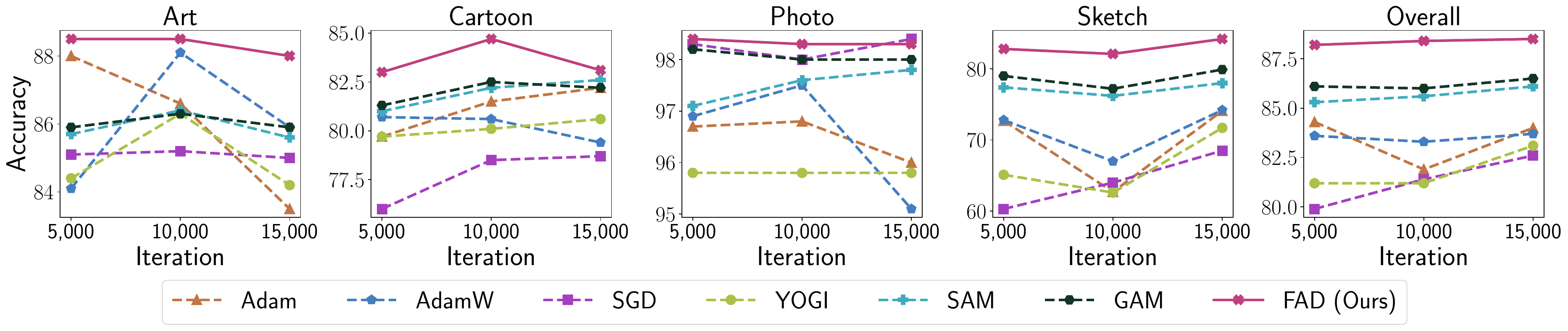}
    \caption{Accuracy of different methods on PACS with different training iterations.}
    \label{fig:iter_pacs}
\end{figure}

\begin{figure}[t]
    \centering
    \includegraphics[width=\linewidth]{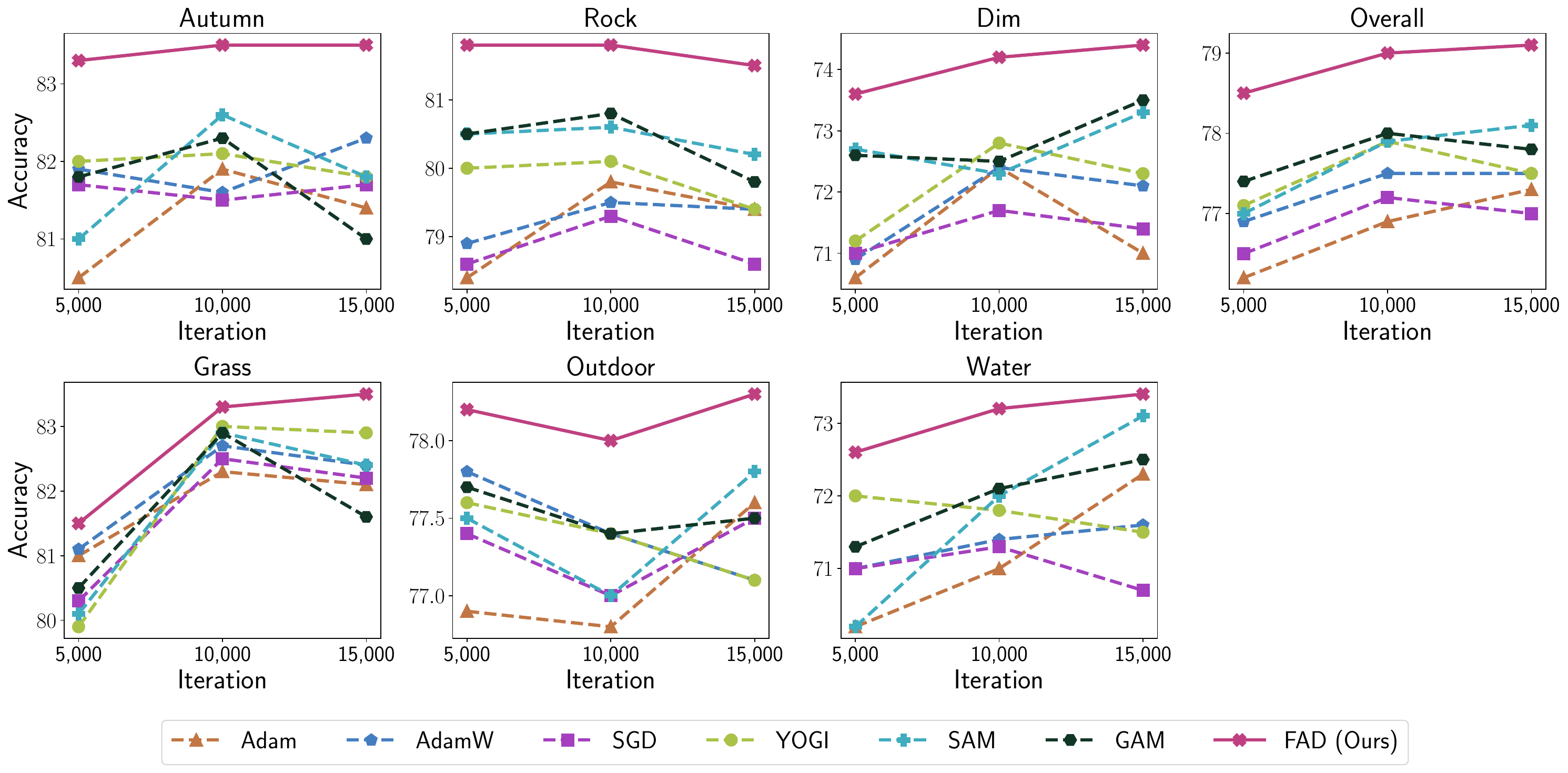}
    \caption{Accuracy of different methods on NICO++ with different training iterations.}
    \label{fig:iter_nico++}
\end{figure}

We show the results of models trained with current optimizers and FAD for various iterations on PACS and NICO++ in Figure \ref{fig:iter_pacs} and Figure \ref{fig:iter_nico++}, respectively. 
\subsubsection{Optimizers with more training iterations on PACS}
We report the detailed results on PACS trained with 10,000 iterations in Table \ref{tab:pacs10k}. FAD consistently outperforms its counterparts on all domains.

\begin{table*}[th]

\centering
\caption{The comparison of optimizers on PACS with 10,000 iterations.}
\begin{tabular}{lcccc|c}
\toprule
\textbf{Algorithm} & art & cartoon & photo & sketch  & \textbf{Avg.} \\
\midrule

Adam  & 86.6 &  81.5 & 96.8 & 62.7 & 81.9\\
AdamW  & 88.1 & 80.6  & 97.5 & 67.0 & 83.3 \\
SGD  & 85.2 & 78.5  & 98.0 & 64.0 & 81.4 \\
YOGI  & 86.3 & 80.1  & 95.8 & 62.6 & 81.2 \\
SAM  & 86.4 & 82.2 & 97.6 & 76.2 & 85.6 \\
GAM  & 86.3 & 82.5  & 98.0 & 77.2 & 86.0 \\

\midrule
FAD (ours) & \textbf{88.5}  & \textbf{84.7}   & \textbf{98.3}& \textbf{82.1}  & \textbf{88.4} \\

\bottomrule
\end{tabular}
\label{tab:pacs10k}
\end{table*}

We report the detailed results on PACS trained with 15,000 iterations in Table \ref{tab:pacs15k}. FAD consistently outperforms its counterparts on all domains.
\begin{table*}[th]
\centering
\caption{The comparison of optimizers on PACS with 15,000 iteration.}
\begin{tabular}{lcccc|c}
\toprule
\textbf{Algorithm} & art & cartoon & photo & sketch  & \textbf{Avg.} \\
\midrule

Adam  & 83.5  & 82.2 & 96.0 & 74.1 & 84.0 \\
AdamW  & 85.9  & 79.4  & 95.1 & 74.2 & 83.7 \\
SGD  & 85.0 & 78.7  & 98.4 & 68.5 & 82.6 \\
YOGI  & 84.2 & 80.6  & 95.8 & 71.7 & 83.1 \\
SAM  & 85.6 & 82.6  & 97.8 & 78.0 & 86.1 \\
GAM  & 85.9 &  82.2 & 98.0 & 79.9 & 86.5 \\

\midrule
FAD (ours) & \textbf{88.0} & \textbf{83.1}   & \textbf{98.3} & \textbf{84.2}  & \textbf{88.5} \\

\bottomrule
\end{tabular}
\label{tab:pacs15k}
\end{table*}

\subsubsection{Optimizers with more training iterations on NICO++}
We report the detailed results on NICO++ trained with 5,000 iterations in Table \ref{tab:nico5k}. FAD consistently outperforms its counterparts on all domains.

\begin{table*}[th]
\centering
\caption{The comparison of optimizers on NICO++. All the models are trained for 5,000 iterations and evaluated following the protocol in DomainBed~\cite{gulrajani2020search}. The best results for each domain are highlighted in bold font.}
\begin{tabular}{lcccccc|c}
\toprule
\textbf{Algorithm} & Autumn & Rock & Dim & Grass  & Outdoor & Water &\textbf{Avg.} \\
\midrule

Adam  & 80.5 & 78.4 & 70.6 & 81.0 & 76.9 & 70.2 & 76.2 \\
AdamW  & 81.9 & 78.9  & 70.9 & 81.1 & 77.8 & 71.0 & 76.9 \\
SGD  & 81.7 & 78.6 & 71.0 & 80.3 & 77.4 & 71.0 & 76.5 \\
YOGI  & 82.0 & 80.0 & 71.2 & 79.9 & 77.6 & 72.0 & 77.1 \\
SAM  & 81.0 & 80.5  & 72.7 & 80.1 & 77.5 & 70.2 & 77.0\\
GAM  & 81.8 & 80.5  & 72.6 & 80.5 & 77.7 & 71.3 & 77.4\\
\midrule
FAD (ours)  &  \textbf{83.3} & \textbf{81.8}   & \textbf{73.6}  & \textbf{81.5}  & \textbf{78.2}  & \textbf{72.6}  &  \textbf{78.5}\\

\bottomrule
\end{tabular}
\label{tab:nico5k}
\end{table*}

We report the detailed results on NICO++ trained with 15,000 iterations in Table \ref{tab:nico15k}. FAD consistently outperforms its counterparts on all domains.

\begin{table*}[th]
\centering
\caption{The comparison of optimizers on NICO++. All the models are trained for 15,000 iterations and evaluated following the protocol in DomainBed~\cite{gulrajani2020search}. The best results for each domain are highlighted in bold font.}
\begin{tabular}{lcccccc|c}
\toprule
\textbf{Algorithm} & Autumn & Rock & Dim & Grass  & Outdoor & Water &\textbf{Avg.} \\
\midrule

Adam  & 81.4 & 79.4 & 71.0 & 82.1  & 77.6 & 72.3  &  77.3\\
AdamW  & 82.3 & 79.4 & 72.1 & 82.4  & 77.1 & 71.6 & 77.5 \\
SGD  & 81.7 & 78.6 & 71.4 & 82.2 & 77.5 & 70.7 & 77.0 \\
YOGI  & 81.8 & 79.4 & 72.3 & 82.9 & 77.1 & 71.5 & 77.5 \\
SAM  & 81.8 & 80.2 & 73.3 & 82.4 & 77.8 & 73.1 & 78.1 \\
GAM  & 82.0 & 79.8 & 73.5 & 81.6 & 77.5 & 72.5 & 77.8\\
\midrule
FAD (ours)  & \textbf{83.5} & \textbf{81.5} & \textbf{74.4}  & \textbf{83.5} & \textbf{78.3} & \textbf{73.4}  & \textbf{79.1} \\

\bottomrule
\end{tabular}
\label{tab:nico15k}
\end{table*}

\subsection{Ablation study}
\begin{figure}[t]
    \centering
    \includegraphics[width=0.9\linewidth]{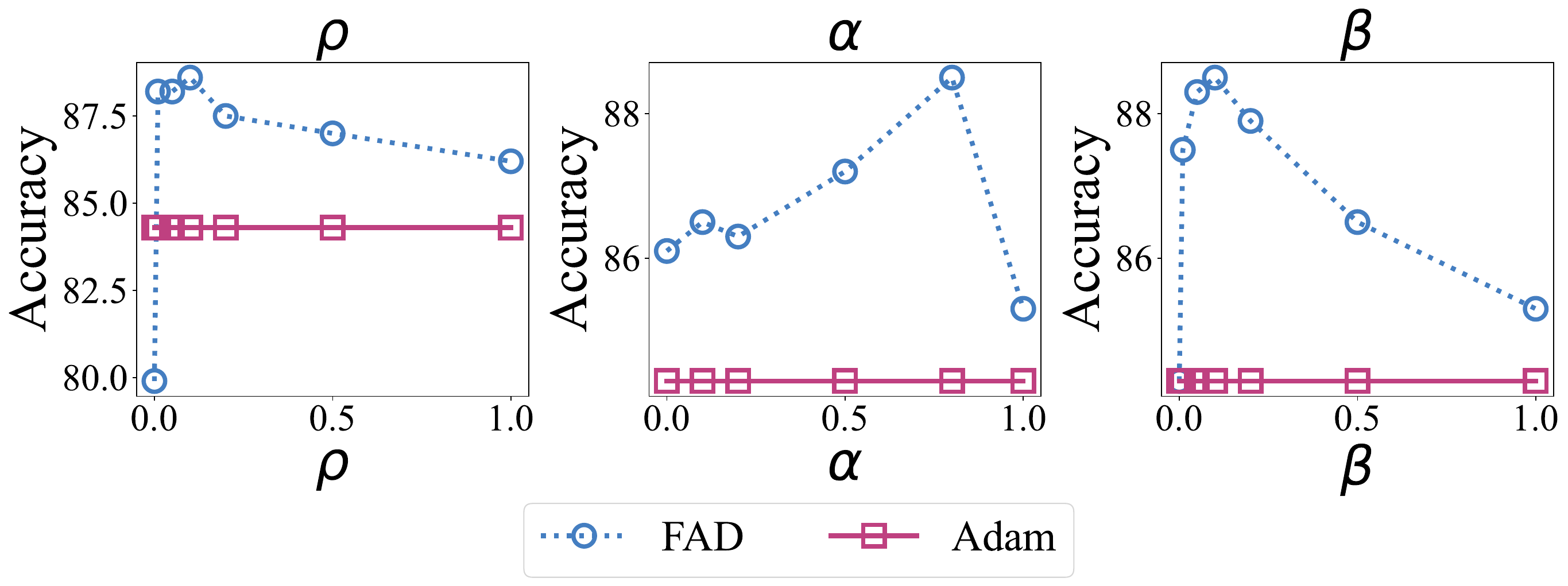}
    \caption{Accuracy of different methods on PACS with different training iterations.}
    \label{fig:ablation}
\end{figure}

FAD has three hyperparameters, namely $\rho$, $\alpha$, and $\beta$. Here we investigate the influence of the choice of them on PACS. The results are shown in Figure \ref{fig:ablation}. 
$\rho$ controls the step length of gradient ascent in FAD. When $\rho$ is set to 0, FAD degenerates into its base optimizer, i.e., SGD. When $\rho$ is larger than 0, FAD consistently outperforms SGD and Adam, as shown in the subfigure on the left of Figure \ref{fig:ablation}. $\alpha$ controls the proportion of intensity distributed between the zeroth-order and first-order flatness. When $\alpha$ is set to 0, FAD degenerates into GAM. Similarly, when $\alpha$ is set to 1, FAD degenerates into SAM. FAD consistently outperforms Adam with various choices of $\alpha$, as shown in the subfigure on the middle of Figure \ref{fig:ablation}. When $\beta$ is set to 0, FAD degenerates into its base optimizer, i.e., SGD. When $\beta$ is larger than 0, FAD consistently outperforms SGD and Adam, as shown in the subfigure on the right of Figure \ref{fig:ablation}. Please note that we only conduct grid search for the ablation study, the hyperparameters in other experiments are chosen based on the validation performance instead of test performance following the protocol in DomainBed. During the training and model selection phase, the information pertaining to the test data is not available.
\end{document}